%% file: main.tex
\documentclass[10pt]{article}
\usepackage[accepted]{tmlr}

\input{math_commands.tex}

\usepackage{hyperref}
\usepackage{url}
\usepackage{amsmath}
\usepackage{amssymb}
\usepackage{mathtools}
\usepackage{booktabs}
\usepackage{multirow}
\usepackage{graphicx}
\usepackage{subcaption}
\usepackage{xcolor}
\usepackage{tcolorbox}
\usepackage{enumitem}
\usepackage{array}

\newcommand{\method}{GRACE}
\newcommand{\methodfull}{Generalist Retrieval-Augmented Correction for Segmentation Errors}
\newcommand{\graceu}{\method-U}
\newcommand{\gracem}{\method-M}

\newcommand{\rev}[1]{{\color{black}#1}}
\newcommand{\revisionfinal}[1]{{\color{black}#1}}

\newcommand{\real}[1]{#1}

\title{Retrieval-Augmented Correction for Generalist Medical \\Image Segmentation}

\author{%
  \name Vincenzo Marcian\`o \email vincenzo.marciano@eurecom.fr \\
  \addr EURECOM \\
  Sorbonne Universit\'e \\
  King's College London
  \AND
  \name Xiaoming Zhang \email xiaoming.zhang@eurecom.fr \\
  \addr EURECOM \\
  Sorbonne Universit\'e
  \AND
  \name Michela Antonelli \email michela.antonelli@kcl.ac.uk \\
  \addr King's College London
  \AND
  \name S\'ebastien Ourselin \email sebastien.ourselin@kcl.ac.uk \\
  \addr King's College London
  \AND
  \name Maria A. Zuluaga \email maria.zuluaga@eurecom.fr \\
  \addr EURECOM \\
  King's College London
}

\def\month{10}
\def\year{2026}
\def\openreview{\url{https://openreview.net/forum?id=PeDqE0SQRe}}

\begin{document}
\maketitle

\begin{abstract}
Automated segmentation is routinely used as the first step in quantitative image-based clinical workflows, where predicted masks are subsequently reviewed and corrected by human operators.
Automated correction of the residual structural errors is currently addressed one anatomy at a time, so the number of trained correctors scales with the number of segmentation targets. We present \method{}, a retrieval-augmented refiner that edits an input mask under context drawn from a shared knowledge base of verified prior cases. The refiner is trained once on a multi-organ corpus spanning compact organs, cavities, branching vessels, and tumour lesions. \rev{In the held-out dataset-transfer evaluation, three held-out datasets query a fixed knowledge base built from training data only. A separate leave-one-anatomy-out experiment tests parameter-free index extension: verified cases of an unseen \revisionfinal{solid} organ enter the knowledge base, and the refiner stays frozen.} \rev{A shape-aware retrieval projector ranks the neighbours. This light projection head sits on a frozen multimodal encoder and maps embedding similarity to the shape agreement of the masks. Thus, the knowledge base returns cases that the refiner can use as a shape prior.} We evaluate \method{} on 18 public datasets spanning CT, MRI and ultrasound, \rev{including three held-out CT/MRI datasets for held-out dataset transfer}, using \rev{nine} refinement configurations and seven upstream segmentors that span weak, mid, and strong operating regimes. \rev{We report each change with the DSC of its input mask. On the weak tier, the input DSC is $0.386$ and \method{} adds up to $+0.35$ DSC. On a strong segmentor, the input DSC is $0.932$ and the mean change is small. Transfer to an imaging modality that is absent from training and from the index is partial and depends on the anatomy. We present it as an open research direction.}
Our results suggest retrieval-conditioned refinement is a practical alternative to anatomy-specific correction models. Open-source code and implementation available at \revisionfinal{\url{https://github.com/robustml-eurecom/grace}}.

\end{abstract}

\section{Introduction}
\label{sec:intro}
Medical image segmentation has advanced rapidly with deep learning~\citep{isensee2021nnu,hatamizadeh2021swin,MedSAM}, becoming the de facto approach for extracting anatomical structures in clinical workflows. Although widely adopted, state-of-the-art models remain prone to structural errors and artifacts that require costly manual correction before clinical use~\citep{azad2023foundational,galati2022accuracy}. These artifacts typically manifest as over-segmentation, under-segmentation, boundary imprecision, or topological violations. Challenging cases involve ambiguous boundaries, small structures, or out-of-distribution pathologies that deviate from the expected normative anatomy~\citep{galati2022accuracy}.

Despite these advances in segmentation accuracy, \emph{automated correction of segmentation errors remains an open problem}. Unlike segmentation, which predicts pixel-wise labels from image intensities, correction requires reasoning about anatomical plausibility and restoring coherent structures from imperfect predictions. Despite growing interest in automated segmentation quality control and refinement ~\citep{valindria2017reverse,robinson2018realtime,audelan2019unsupervised,liu2019alarm,Robinson2019Automated,larrazabal2020post,painchaud2020,wang2020deep,fournel2021medical,lin2022novel,qiu2023qcresunet,marciano2026diffusion}, the field currently lacks \revisionfinal{generalisable}, fully-automated solutions.

Existing approaches largely fall into two restrictive paradigms. On the one hand, single-anatomy specialists~\citep{larrazabal2020post,wang2020deep,painchaud2020} learn to encode anatomical shape priors, which are then used to validate and improve anatomically implausible segmentations. However, while effective, these approaches require training a dedicated model from scratch for each new organ or imaging modality. Another set of techniques interprets refinement as a signal restoration problem, where given a corrupted mask, the goal is to recover the anatomically plausible structure~\citep{yuan2020segfix,terris2025ram}. While such models demonstrate strong capabilities in generic signal restoration, they do not explicitly incorporate structured anatomical priors, which are critical for correcting geometric segmentation errors.



We argue that medical segmentation refinement can be naturally formulated as a \emph{Retrieval-Augmented Generation (RAG)} problem~\citep{lewis2020retrieval}. Given a corrupted segmentation, a model retrieves anatomically similar 
cases from a knowledge base and conditions the refinement process on this retrieved context. In this way, anatomy-specific information is provided at inference time rather than being encoded into the model parameters, enabling a single refiner to operate across organs and imaging modalities represented in the knowledge base. Although retrieval has become a powerful paradigm for knowledge-intensive learning, its potential for anatomy-agnostic segmentation refinement remains largely unexplored.
%
To this end, we present \method{} \revisionfinal{(\methodfull{})} as a retrieval-conditioned segmentation-refinement framework (Fig.~\ref{fig:pipeline}) that separates anatomy-aware retrieval from the refiner representation. The framework itself does not commit to a particular refinement representation: pixel-space and latent-space refiners are treated as interchangeable instantiations of a common retrieval interface. Our contributions are, thus, summarised as follows:

\begin{figure}[!t]
    \centering
    \includegraphics[width=\linewidth]{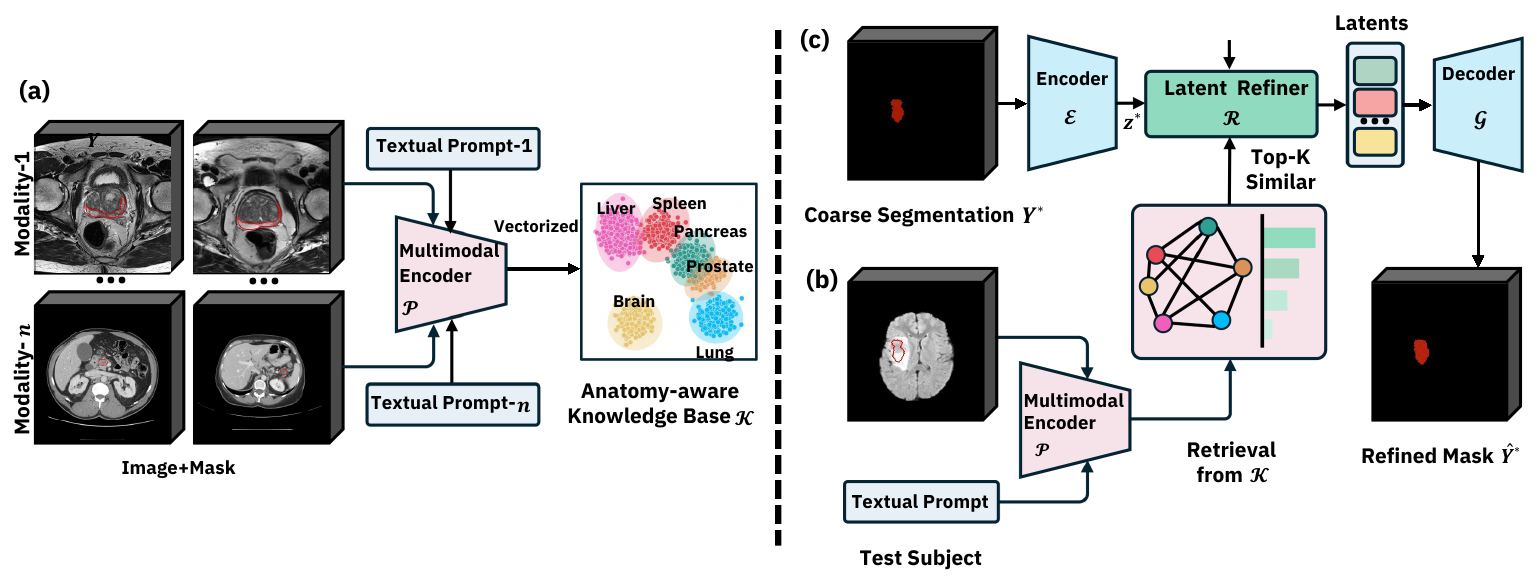}
    \caption{\textbf{Overview of \method{}.} (a) The knowledge base stores multimodal embeddings of verified cases together with their corresponding masks. (b) At inference, the query (corrupted mask $Y^*$ and image $X$) is embedded, and the top-$k$ same-class neighbours are retrieved from a FAISS index. The top-1 mask $Y_r$ is used as an exemplar. (c) The refiner produces the corrected mask $\hat Y^*$ from the query and retrieved context using either a pixel-space or a latent-space instantiation. Figure elements that depict a compact latent representation illustrate the latent variant; the pixel-space variant follows the same retrieval interface without an encoder--decoder pair. \revisionfinal{The multimodal encoder $\mathcal{P}$ is frozen, and the projection head $\phi_\psi$ (Sec.~\ref{sec:proj}) is not drawn. In (c), $z^*{=}\mathcal{E}(Y^*)$.}}
    \label{fig:pipeline}
\end{figure}

\begin{itemize}[leftmargin=*,topsep=2pt,itemsep=1pt]
    \item A retrieval-augmented framework for segmentation refinement whose common interface supports both pixel-space and latent-space refiners. The retrieval step supplies anatomy-specific information at inference time, allowing a single anatomy-agnostic refiner to generalise across \revisionfinal{the} organs and imaging modalities \revisionfinal{represented in the knowledge base}. The framework can use a fixed knowledge base or incorporate verified support cases without retraining the refiner.
    
    \item \rev{Exemplar-conditioned refinement. The retrieved mask enters the refiner as a spatial channel and gives the shape prior for the correction. When we set this channel to zero, the refiner loses $40\%$ of its gain at the most severe preset (Sec.~\ref{sec:retanalysis}). A shape-aware retrieval projector on a frozen multimodal encoder selects the same-class case. Its contribution is smaller, and we report it as a component ablation (Table~\ref{tab:retrieval_projection}).}
    
    \item Three complementary instantiations of the proposed framework spanning deterministic pixel-space refinement, deterministic latent refinement, and latent flow matching. \rev{We compare all three across the upstream operating regimes and on the trade-off between accuracy and runtime.}
    
    \item Extensive validation across 18 public datasets, multiple imaging modalities, seven upstream segmentors, and \rev{nine} refinement configurations, \rev{including three held-out datasets for held-out dataset transfer and a leave-one-modality-out study. We also analyse retrieval, knowledge-base scaling, robustness to a wrong class query, case-level failure rates and statistical significance}, and release the code, pretrained models, and knowledge base.
\end{itemize}

\section{Related Work}
\label{sec:related}

\paragraph{Segmentation quality control and refinement.}
Segmentation refinement is a distinct problem from segmentation itself and has been approached along three main lines. The first is classical, hand-designed post-processing that enforces geometric or topological constraints on the raw prediction: morphological operations and connected-component filtering to remove small false-positive blobs or retain the largest anatomically plausible regions~\citep{isensee2021nnu,Dulau2023} and topological penalties ensuring continuous boundaries \citep{Hu2019,Li2025}. These methods are fast and interpretable but cannot recover structure that was never in the raw prediction, and their handcrafted rules do not transfer across anatomies. The second line replaces handcrafted rules with 
learned anatomical priors, typically encoded through autoencoders, variational autoencoders, or more recently latent diffusion models~\citep{Oktay2017,larrazabal2020post,wang2020deep,painchaud2020,galati2021efficient,marciano2026diffusion}. These methods learn a manifold of anatomically plausible segmentations that guides refinement. However, as the manifold is typically learned for a single anatomy, these methods require retraining for each new organ or imaging modality.
Finally, the third line does not consider a dedicated corrector, approaching the problem via test-time adaptation (TTA)~\citep{karani2021test}. 
Instead, the upstream segmentor is briefly refitted on each test volume to close a domain gap. TTA improves segmentation performance without producing a corrected mask directly. Therefore, it is complementary to a refiner rather than a substitute for it. \method{} sits in the refiner slot, replaces the per-anatomy autoencoder with a single retrieval-conditioned refiner trained once on multi-organ data, and preserves the conservative \revisionfinal{behaviour} of morphological post-processing on inputs that do not need correction.

\paragraph{Foundation models in medical imaging.}
Recent foundation and generalist models have broadened anatomical coverage through diverse paradigms, including prompt-based segmentation~\citep{MedSAM,wang2025sam,ma2025medsam2,wong2024scribbleprompt}, large-scale supervised learning~\citep{wasserthal2023totalsegmentator}, few-shot segmentation~\citep{butoi2023universeg}, and text-conditioned segmentation~\citep{rokuss2025voxtell}. While these methods demonstrate broad \revisionfinal{generalisation} across anatomies and modalities, they are designed for forward segmentation prediction rather than post-hoc correction and refinement. RAM~\citep{terris2025ram} extends the foundation-model paradigm to generic inverse imaging problems without task-specific training,  but operates in pixel space and does not explicitly exploit anatomical context for segmentation correction.


\paragraph{Retrieval-augmented learning in medical imaging.}
Retrieval-augmented learning~\citep{lewis2020retrieval} has recently been explored in medical AI for report generation~\citep{xia2024mmed,xiong2024benchmarking,hsu2025medplan} and few-shot segmentation from retrieved annotated examples~\citep{ahmed2025j,zhao2025retrieval}. In these settings, retrieval provides external knowledge to improve prediction or support inference. In contrast, retrieval for direct segmentation refinement has received little attention. \method{} retrieves anatomically similar cases to provide inference-time context for correcting an existing segmentation.


\paragraph{Latent-space exploration in medical imaging.}
Compact latent representations have become a standard paradigm for \revisionfinal{modelling} anatomically plausible structures in medical imaging.
Early formulations use variational autoencoders (VAEs) as shape priors, learning latent manifolds of plausible organ shapes for segmentation refinement and shape reconstruction~\citep{painchaud2020,larrazabal2020post,beetz2022meshvae}.
 Vector-Quantised VAEs~\citep{razavi2019generating} and multi-scale volumetric encoders~\citep{pinaya2022brain,fernandez2024generating} extended this framework to high-resolution 3D image and mask generation.
%
Latent diffusion models~\citep{rombach2021highresolution} and, more recently, rectified flow models~\citep{lipman2022flow} enable efficient generative sampling directly in these compact latent spaces, making full-volume 3D synthesis computationally tractable~\citep{guo2025maisi,chen20243dmeddiffusion,lin2024medlsdm}. Conditioning strategies have subsequently evolved from partial-noise editing, such as SDEdit~\citep{meng2022sdedit}, to retrieval-guided generation, where retrieved exemplars provide structural guidance during sampling~\citep{fang2024raag}. \method{} builds on this latter paradigm by performing class-filtered, shape-aware retrieval to guide segmentation refinement, and supports deterministic pixel-space, deterministic latent-space, and latent flow-based refiners within a unified framework.


\section{Method}
\label{sec:method}

\subsection{Problem formulation}
\label{sec:framework}
Let $Y^* \in \{0,1\}^{D \times H \times W}$ be a mask produced by an upstream segmentation model on image $X$, and let $Y$ be the (unknown) ground-truth mask. The refinement problem is to produce $\hat Y^* \approx Y$ from $(Y^*, X)$ and a knowledge base $\mathcal{K}$ of previously verified cases. The framework decomposes this task into three components:
\begin{enumerate}[leftmargin=*,topsep=1pt,itemsep=1pt]
    \item A knowledge base $\mathcal{K} = \{(e_i, Y_i, X_i, \text{class}_i)\}_{i=1}^{N}$ that stores each verified case together with a multimodal embedding $e_i$ that supports semantic retrieval (Sec.~\ref{sec:kb}).
    \item A retrieval step that maps the query $(Y^*, X)$ to a raw multimodal embedding $e^* = \mathcal{P}(Y^*, X)$ and then to the indexed representation $\revisionfinal{q^*}=\phi_\psi(e^*)$, and returns a context set
    \begin{equation*}
        c \;=\; \big(\{e_{r_1}, \ldots, e_{r_k}\},\; Y_r\big),
    \end{equation*}
    where $\revisionfinal{q_i}=\phi_\psi(e_i)$ and $\{r_1,\ldots,r_k\}=\mathrm{NN}_k(\revisionfinal{q^*};\{\revisionfinal{q_i}:\mathrm{class}_i=\mathrm{class}^*\})$ are the top-$k$ cases for the query's target class $\mathrm{class}^*$. In raw-space ablations, $\phi_\psi$ is replaced by the identity map. $Y_r$ is the ground-truth mask of the top-ranked retrieved case.
    \item A refiner $\mathcal{R}_\theta$ that maps the query and the retrieved context to a corrected mask,
    \begin{equation}
        \hat Y^* \;=\; \mathcal{R}_\theta\!\left(Y^*,\, X,\, Y_r,\, \{e_{r_j}\}_{j=1}^{k}\right).
        \label{eq:refiner}
    \end{equation}
\end{enumerate}
The retrieval step supplies anatomy-specific information at inference time; $\mathcal{R}_\theta$ is trained once on multi-organ data and is anatomy-agnostic at parameter level.
Datasets containing anatomies represented in the training corpus can query the fixed knowledge base directly. When verified masks are available, disjoint support cases can instead be added to $\mathcal{K}$ without retraining $\mathcal{R}_\theta$; Sec.~\ref{sec:leave_one_out} evaluates this optional parameter-free index extension separately from held-out dataset transfer.

\subsection{Anatomy-aware knowledge base}
\label{sec:kb}
\begin{figure}
    \centering
    \includegraphics[width=\linewidth]{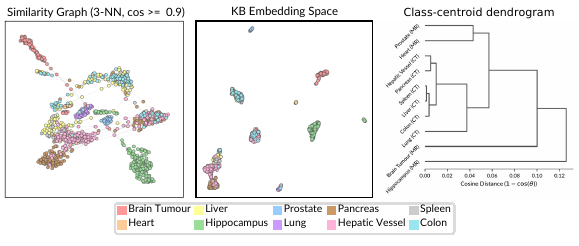}
    \caption{\textbf{Semantic topology and latent anatomical hierarchy of the anatomy-aware knowledge base $\mathcal{K}$}. (Left) Similarity graph: A 3-nearest \revisionfinal{neighbour} connectivity graph where edges denote a cosine similarity $\ge 0.9$, illustrating strong intra-class cohesion. (Middle) Embedding space projected on a t-SNE vector space. (Right) Semantic hierarchical clustering: A dendrogram of class centroids computed via average linkage and cosine distance. The topological structure demonstrates that $\mathcal{K}$ intrinsically captures underlying anatomical relationships and modality-specific groupings (e.g., abdominal CT and neurological MR clusters).}
    \label{fig:kg}
\end{figure}

For each verified case, the binary mask is overlaid as a contour on each corresponding slice of the paired scan, aggregated in a 3D volume as suggested in \citet{sellergren2025medgemma}, and the composite is paired with a structured textual prompt:
\begin{tcolorbox}[colback=black!5!white, colframe=black!25!white, arc=2mm, boxrule=0.5pt, left=3pt, right=3pt, top=3pt, bottom=3pt, fontupper=\ttfamily\small]
You are an expert radiologist. The highlighted region (red contours) represents: \textcolor{purple}{\{class\_name\}}. Describe its visual characteristics.
\end{tcolorbox}
\noindent The composite is processed by a frozen multimodal \revisionfinal{encoder} $\mathcal{P}$. Unless otherwise stated, we use
MedGemma 4B~\citep{sellergren2025medgemma} to obtain raw embeddings $e_i \in \mathbb{R}^{2560}$. The main pipeline maps these embeddings to 256-dimensional shape-aware vectors using $\phi_\psi$ (Sec.~\ref{sec:proj}); the projected vectors are L2-normalised and indexed with FAISS~\citep{douze2025faiss}. Raw-space indexing is used only in the corresponding ablations. \rev{FAISS is a similarity-search library. It keeps the normalised embeddings in a flat inner-product index. Because the vectors have unit length, the inner-product search returns the cosine nearest neighbours. The index makes the top-$k$ search fast enough for use at inference.}
Figure~\ref{fig:kg} illustrates the resulting embedding topology via a $k$-nearest-neighbour connectivity graph and hierarchical clustering, revealing strong intrinsic modality alignment and an emergent anatomical hierarchy within the embedding space. Extending $\mathcal{K}$ with a verified support case requires only embedding and indexing that case, not retraining the refiner.


\paragraph{Construction protocol.}
In the main experiments and the held-out-dataset evaluation, cases enter the knowledge base only from the training splits listed in Sec.~\ref{sec:setup}; validation cases, test cases, and every case from the three held-out datasets in Sec.~\ref{sec:unseen} are excluded. The leave-one-anatomy-out support experiment (Sec.~\ref{sec:leave_one_out}) and the optional longitudinal-memory analysis (Appendix~\ref{sec:supp_longitudinal}) are explicitly identified extensions of this fixed-index protocol. Their indexed support cases are disjoint from the corresponding evaluation queries. Multi-class masks are decomposed into per-class binary instances, \revisionfinal{and each indexed instance is one entry}. 
A same-class index is materialised at indexing time, avoiding class filtering over the full database during inference. No scored evaluation query is indexed as its own retrieval candidate. In the reported configuration, $\mathcal{K}$ contains $N{=}3{,}913$ per-class instances drawn from the training splits of the 15 training datasets (Table~\ref{tab:dataset_scope})\revisionfinal{, out of the 11,900 non-empty instances of these splits (Appendix~\ref{sec:supp_datasets})}.

\subsection{Shape-aware retrieval projector}
\label{sec:proj}
\rev{$\mathcal{P}$ denotes the frozen MedGemma encoder only. It maps a pair of composite image and prompt to a raw 2560-dimensional embedding. The L2 normalisation, the projector $\phi_\psi$ and the FAISS index are separate stages that operate on the output of $\mathcal{P}$. Cosine similarity on the raw embeddings orders the neighbours by general multimodal appearance. The refiner uses a same-class mask as a spatial exemplar, so it needs neighbours with a similar shape.} We fit a projector $\phi_\psi:\mathbb{R}^d\to\mathbb{R}^{d'}$ on $\mathcal{K}$ so that cosine similarity in the projected space approximates shape-Dice between the underlying masks:
\begin{equation}
    \mathcal{L}_{\phi} = \frac{1}{|\mathcal{B}|} \sum_{(i,j)\in\mathcal{B}} \Big( \langle \phi_\psi(e_i), \phi_\psi(e_j)\rangle - \mathrm{Dice}\big(\widetilde m_i, \widetilde m_j\big) \Big)^2,
    \label{eq:shape_projector}
\end{equation}
where $\widetilde m$ denotes a mask cropped to its bounding box and resampled to a shared $32^3$ cube so that position and scale are removed and only shape is preserved. $\phi_\psi$ is a 3-layer MLP (hidden width 512, output $d'{=}256$) trained for 20 epochs on uniformly-sampled node pairs from $\mathcal{K}$. 
The projected vectors are L2-normalised, so the inner product in Eq.~\ref{eq:shape_projector} is their cosine similarity.
The projector is fitted after the raw knowledge-base embeddings have been constructed. Its normalised outputs replace the raw vectors as FAISS keys, leaving the refiner and its training unchanged.

\subsection{Refiner interface}
\label{sec:refiner_interface}
The retrieved $Y_r$ enters $\mathcal{R}_\theta$ as a spatial conditioning channel that supplies a shape prior, while the sequence $\{e_{r_j}\}_{j=1}^{k}$ provides semantic context through cross-attention, whenever supported by the underlying architecture. 
This interface is compatible with both supervised feedforward networks trained on pixel-wise labels and conditional generative models that sample from $p(Y \mid Y^*, X, Y_r, \{e_{r_j}\}_{j=1}^{k})$. The refiner may operate directly in pixel space or in a compact latent space obtained through a mask autoencoder. In the latter case, the retrieved $Y_r$ is projected through the same encoder. The latent refinement interface is independent of the particular autoencoder architecture; throughout this work, the latent variants use the frozen encoder of the MAISI VAE~\citep{guo2025maisi} together with a mask-tuned decoder. Any compatible mask autoencoder can be substituted without modifying the refinement procedure.

These two factors, the refinement paradigm (deterministic or generative) and the representation (pixel or latent space), can be combined to instantiate different refinement architectures within the same retrieval framework. In our work, they give rise to three variants: a deterministic pixel-space network (\graceu{}), a deterministic latent autoencoder (\gracem{}), and a latent flow-matching model (\gracem-D{}).
\rev{\graceu{} is a 3D U-Net with three input channels: the corrupted mask, the image and the retrieved exemplar. It gives a foreground logit for each voxel in one forward pass. \gracem{} encodes the corrupted mask and the exemplar into a $4{\times}24^3$ latent with a frozen encoder. A small module fuses the two latents, and a mask-tuned decoder gives the refined mask. \gracem-D{} uses the same latent interface but integrates a learned velocity field with an ODE solver. Thus, several runs on one input give a set of plausible refinements. All three variants use the same retrieval context. Appendix~\ref{sec:refiner_paradigms} gives the full architectures.}


\subsection{Two-phase training and corruption schedule}
\label{sec:refiner_train}
Training $\mathcal{R}_\theta$ directly on corrupted masks produced by one specific upstream segmentor risks injection bias: the refiner overfits to the failure modes of that segmentor. \rev{To prevent this bias, we train the refiner on synthetic corruptions of the ground-truth masks. A corruption combines morphological erosion and dilation, elastic boundary deformation, boundary noise, false-positive blobs and axis-aligned slice dropout~\citep{jebril2025shape}.} Three severity presets (\emph{default}, \emph{mid}, \emph{strong}) define the synthetic-corruption experiments; a fourth training preset (\emph{mixed}) samples uniformly across the three severities. \rev{Two more presets, \emph{very\_strong} and \emph{extreme}, go \revisionfinal{beyond} the range of the training severities. We use them only for evaluation, in the analyses of Sec.~\ref{sec:retanalysis}.} Representative synthetic corruptions are shown in Fig.~\ref{fig:corruption_examples}.

\rev{Training has two phases. Phase~1 trains the refiner, and Phase~2 fits the retrieval projector $\phi_\psi$ of Sec.~\ref{sec:proj}. In Phase~1, we train the refiner from random initialisation for 100 epochs at $96^3$ resolution. We use AdamW with a learning rate of $10^{-4}$, a cosine schedule, batch size 8, gradient clipping at 1.0 and mixed precision. The loss combines Dice and cross-entropy against the clean ground truth. The \emph{mixed} preset corrupts each input with probability $0.95$, so a small fraction of clean inputs keeps the identity behaviour. The exemplar is set to zero with probability $0.10$. Thus, the refiner stays usable when retrieval gives no result, and it cannot use the exemplar as a shortcut. The corruption changes for each sample and each epoch. In Phase~2, we fit $\phi_\psi$ for 20 epochs on pairs of knowledge-base cases (Eq.~\ref{eq:shape_projector}). Phase~2 does not change the refiner.}

\section{Experiments}
\label{sec:experiments}

\subsection{Experimental setup}
\label{sec:setup}

\paragraph{Datasets.}
The training corpus covers 15 public datasets across CT, MRI, and US and contains \revisionfinal{3,866} volumes over 38 anatomical structures across the abdomen, thorax, pelvis, brain, and cardiac chambers (Table~\ref{tab:dataset_scope} in Appendix~\ref{sec:supp_datasets}). Multi-class labels are converted to per-organ binary instances; for multi-channel Medical Segmentation Decathlon (MSD)~\citep{antonelli2022medical} tasks the \revisionfinal{T2-weighted} modality is used. Three datasets are excluded from both training and indexing and are used for held-out dataset-transfer evaluation: ProstateX~\citep{armato2018prostatex} (346 subjects, multiparametric prostate MRI), CHAOS~\citep{CHAOS2021} (40 CT subjects, liver / kidneys / spleen; CHAOS also has MRI scans of these organs, which Sec.~\ref{sec:unseen_modality} uses), and LUMIERE~\citep{suter2022lumiere} (91 GBM patients across 2--20 longitudinal visits each, 599 timepoints).

\paragraph{Metrics.} We report Dice coefficient (DSC, $\uparrow$), 95th-percentile Hausdorff distance (HD95, $\downarrow$), and signed deltas over the raw upstream prediction. Unless a caption states otherwise, ``$\pm$'' denotes the sample standard deviation across cases. Figure~\ref{fig:refiner_performance} instead shows variability across anatomy--dataset pairs, and Table~\ref{tab:det_vs_gen} reports runtime variability across repeated runs. HD95 is reported in millimetres for real-segmentor evaluations and in voxels for synthetic-corruption analyses.

\paragraph{Upstream segmentors.} Seven segmentors span three operating regimes. \emph{Weak}: a MONAI\revisionfinal{~\citep{cardoso2022monai}} 3D U-Net trained for 30 epochs on 10\% of each organ split (U-Net@30). \emph{Mid}: MedSAM2~\citep{ma2025medsam2} and ScribblePrompt~\citep{wong2024scribbleprompt} used with an oracle bounding box derived from the ground truth on the highest-coverage axial slice, and SwinUNETR-BTCV~\citep{hatamizadeh2021swin}. ScribblePrompt and SwinUNETR-BTCV are used for the held-out dataset evaluation (Sec. \ref{sec:unseen}). \emph{Strong}: TotalSegmentator~\citep{wasserthal2023totalsegmentator} in fast mode on CT organs; a MONAI 3D U-Net trained for 150 epochs on the full split (U-Net@150); a MONAI SegResNet trained for 150 epochs on the full split (SegResNet). This split characterises \method{} when the input carries little usable signal as well as when it is close to correct.

\rev{\paragraph{What the refiner is trained on.} We split each dataset case-wise into disjoint training and test parts. (i) For Table~\ref{tab:universal} and the synthetic analyses, one refiner trains on the training split, and we evaluate it on the test split. It sees other cases of the same datasets, but never the evaluated cases. (ii) For Table~\ref{tab:unseen}, we apply the same refiner to three datasets that are absent from training and from the index. (iii) For Table~\ref{tab:transfer} (Block A), we retrain the refiner from scratch for each held-out anatomy. That anatomy is absent from training and from the index. At test time, only support cases enter the index, and they are disjoint from the queries. (iv) Block B of Table~\ref{tab:transfer} applies the same protocol to a full modality. We rebuild the index for each protocol.}

\paragraph{Refiner baselines.} \rev{Nine} refiner variants are compared: (i) \emph{no refinement} (raw upstream); (ii) \emph{morphological} (binary closing then binary opening with a $3{\times}3{\times}3$ structuring element); (iii) \emph{test-time augmentation (TTAug)} with eight views and mean aggregation; (iv) PostDAE~\citep{larrazabal2020post}; (v) RAM~\citep{terris2025ram}; (vi) a MONAI 3D U-Net refiner without retrieval (\emph{U-Net w/o RAG}) that takes $(Y^*, X)$ as input; (vii) the same U-Net with the retrieved exemplar as a third input channel (\graceu{}); (viii) a latent autoencoder refiner that fuses the exemplar in the latent bottleneck before decoding (\gracem{}); \rev{(ix) the same latent refiner with flow-matching inference (\gracem-D{}), averaged over three ODE seeds}.
\rev{No baseline receives the target class as input. PostDAE uses the class implicitly, through one model for each anatomy. This is the strongest form of class conditioning. The morphological, TTAug and RAM baselines cannot use a label. U-Net w/o RAG receives only $(Y^*,X)$. \method{} uses the class to select the retrieval candidates. To separate the effect of the class from the effect of the exemplar, we also train U-Net w/o RAG with the class label. A one-hot embedding and FiLM inject the label, and no exemplar is given. The label improves only the weak inputs \revisionfinal{($+0.081$ vs.\ $+0.049$ on U-Net@30)} and does not change the mid and strong inputs \revisionfinal{($-0.002$ on MedSAM2, $-0.004$ on TotalSegmentator)}. \graceu{} still gives a larger gain on weak inputs. Thus, the class label explains little of the advantage of \method{}. This baseline is different from the zeroed-exemplar condition of Sec.~\ref{sec:retanalysis}, which evaluates a trained \method{} model with an empty retrieval channel.} \rev{Appendix~\ref{sec:refiner_paradigms} gives the details of the three \method{} variants. The U-Net w/o RAG and the three \method{} variants use the same corruption schedule (Sec.~\ref{sec:refiner_train}). Only the \method{} variants use the retrieval pipeline of Sec.~\ref{sec:kb}.}

\subsection{\method{} as a universal refiner}
\label{sec:universal}
\begin{figure}[!t]
    \centering
    \includegraphics[width=\linewidth]{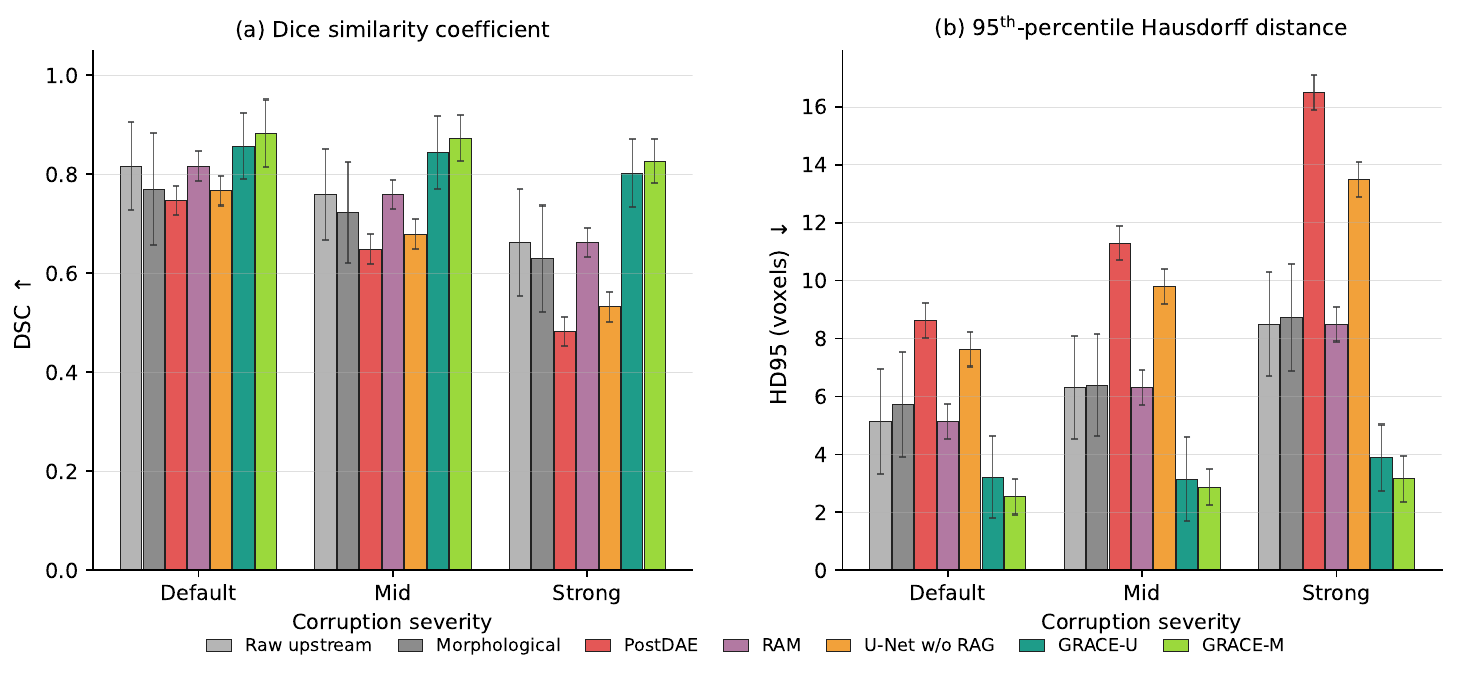}
    \caption{\textbf{Refiner performance across corruption severities.} Bars show mean DSC (a) and mean HD95 (b) across seven anatomy--dataset pairs (Liver, Pancreas, Spleen from MSD; Liver, Pancreas, Spleen from AbdomenCT1K; Spleen from AMOS22); error bars show the sample standard deviation across those seven pair-level values. Seven refiners are compared per severity group: raw upstream (grey), morphological (dark grey), PostDAE (red), RAM (purple), U-Net w/o RAG (amber), \method-U{} (blue-green), and \method-M{} (yellow-green). TTAug requires repeated upstream inference and is therefore not defined for the precomputed synthetic-corruption inputs in this figure.}
    \label{fig:refiner_performance}
\end{figure}

Table~\ref{tab:universal} reports DSC and HD95 on real upstream segmentor outputs; Fig.~\ref{fig:refiner_performance} summarises the evaluation under the three synthetic corruption severities. \method-M{} yields the largest DSC lift in three of the five upstream settings, with \method{}-U ahead on U-Net@30 and U-Net@150, and the \revisionfinal{three \method{} variants provide the three} strongest HD95 reductions in every setting. \rev{Appendix~\ref{sec:failure} gives how often refinement decreases the DSC for each band of input DSC.}

\input{tables/universal.tex}

The benefit of retrieval grows monotonically with corruption severity, and \revisionfinal{\graceu{} and \gracem{}} reduce HD95 at every severity while retrieval-free refiners are less stable (PostDAE and U-Net w/o RAG both degrade under stronger corruption). The same pattern holds for real upstream segmentors in Table~\ref{tab:universal}: strong-tier mean changes remain non-negative for \revisionfinal{all three variants} and are much smaller than the weak-tier gains.

\rev{\paragraph{PostDAE.} PostDAE decreases the DSC on every tier. The largest loss is on TotalSegmentator ($-0.487$ DSC, $+10.2$~mm HD95). PostDAE replaces the input with its reconstruction through a fixed latent prior. Thus, it projects each mask onto its learned manifold. On an almost correct input, this projection removes correct detail. This explains the largest loss on the strongest upstream. PostDAE was also designed for 2D chest X-ray masks. We apply it slice by slice to 3D abdominal organs, and nothing keeps the slices consistent. This effect is largest on HD95. The row thus shows the effect of a single-anatomy 2D shape prior on volumetric multi-organ data. This is the gap that our work addresses.}

\paragraph{Comparison of the variants.} \revisionfinal{The three variants} achieve comparable aggregate performance but differ in their behaviour on severely degraded inputs. Consistent with Table~\ref{tab:universal}, \method-M{} produces the larger mean lift on the mid and most of strong tiers, while \revisionfinal{all three variants} have small non-negative mean changes on \revisionfinal{the strongest segmentor, TotalSegmentator}. Mechanistically, \method-U{} is a single-pass feedforward network and produces more conservative corrections when little reliable signal is available; \method-M{} refines in a frozen latent space and can synthesise anatomically plausible completions guided by the retrieved exemplar. \rev{\gracem-D{} stays within $0.01$ DSC of \gracem{} and needs about seven times more compute (Table~\ref{tab:det_vs_gen}). Its value is the uncertainty estimate for each sample, and it does not increase the accuracy.}

\subsection{Generalisation to held-out datasets}
\label{sec:unseen}
We evaluate \method{} on previously unseen datasets using real upstream segmentors.
Three held-out datasets are used: ProstateX, CHAOS, and LUMIERE. No case from these datasets is added to $\mathcal{K}$ for this evaluation: each query retrieves only from the fixed training-corpus index described in Sec.~\ref{sec:kb}, and no model parameter is updated. \rev{We call this protocol held-out dataset transfer. The anatomies and modalities of these datasets also occur in other training datasets. Thus, the protocol measures transfer under dataset and acquisition shift, and it is not a zero-shot evaluation. Sec.~\ref{sec:unseen_modality} removes a full modality.} When verified target-dataset masks are inserted into $\mathcal{K}$ instead, the protocol becomes a parameter-free index extension as in Sec.~\ref{sec:leave_one_out}.
Table~\ref{tab:unseen} reports raw and refined DSC and HD95 for \revisionfinal{three upstream segmentors (one or two per dataset)}. Both \method{} variants improve DSC on every (segmentor, dataset) pair, and \graceu{} matches or exceeds \gracem{} on two scenarios, with \gracem{} ahead on the CHAOS $\times$ TotalSeg and LUMIERE $\times$ ScribblePrompt. The ordering is therefore dataset- and segmentor-dependent, unlike Table~\ref{tab:universal}, where \gracem{} \revisionfinal{led in three of the five settings}. A possible explanation is that the mask-tuned decoder of \gracem{} is fitted to the training-corpus mask distribution, so its advantage narrows under dataset shift, whereas the pixel-space \graceu{} is less tied to that distribution. Appendix~\ref{sec:supp_longitudinal} reports a separate longitudinal retrieval analysis on LUMIERE.

\input{tables/unseen.tex}

\subsection{Transfer to a held-out anatomy or modality}
\label{sec:leave_one_out}
\label{sec:unseen_modality}
The held-out datasets of Sec.~\ref{sec:unseen} have anatomies and modalities that also occur in the training corpus. This subsection removes a full anatomy or a full imaging modality. In both experiments, we remove the held-out data from the training data and from the retrieval index. Then we retrain the refiner from scratch. Table~\ref{tab:transfer} gives the results.

\input{tables/transfer.tex}

\paragraph{Held-out anatomy.} We evaluate whether \method{} can refine a previously unseen anatomical class without updating its parameters. For each anatomy $a$ in \{Liver, Spleen, Pancreas, Prostate PZ\}, we retrain the refiner without $a$. At test time, we add verified support cases of $a$ to $\mathcal{K}$. Thus, the experiment is a parameter-free index extension. The refiner receives no gradient from $a$, but the index gives it examples of $a$. The support cases are disjoint from the evaluation queries. Both variants improve every held-out anatomy, with mean gains of $+0.202$ DSC for \graceu{} and $+0.203$ for \gracem{} (Table~\ref{tab:transfer}, Block A). The inputs are weak, with a raw DSC from $0.013$ to $0.421$. Each organ needs $118$ to $302$ verified masks, and all four are \revisionfinal{solid organs or organ zones}. The experiment shows that an index extension alone improves a held-out \revisionfinal{solid} organ without a parameter update. It does not show zero-shot generalisation to unseen anatomies.

\rev{\paragraph{Held-out modality.} For each modality in \{US, MRI, CT\}, we remove all its datasets. No case of that modality enters the index at test time. Ultrasound is the clearest case. CAMUS has the same three cardiac structures as ACDC, so the anatomy stays almost constant. The imaging physics, the speckle and the sector geometry change. The MRI holdout is more difficult. Some MRI anatomies, such as the hippocampal subregions, do not occur in the CT and ultrasound data. Thus, the modality and part of the anatomy are both unseen.}

\rev{Refinement transfers from CT to MRI, but it loses most of its in-distribution gain (Table~\ref{tab:transfer}, Block B). An MRI-free \revisionfinal{\graceu{}} refiner with an index of 713 CT livers increases the DSC of CHAOS MRI liver from $0.515$ to \revisionfinal{$0.599$} ($+0.084$). \revisionfinal{On the same CHAOS MRI queries, the refiner trained with MRI gains $+0.349$.} Refinement does not transfer to ultrasound. At an input DSC of $0.438$, the US-free refiner decreases the DSC to $0.236$. The refiner trained with ultrasound reaches $0.772$. Retrieval still selects the correct structure for all 101 CAMUS queries. However, the shape agreement between a query and its retrieved pool drops to $29\%$ of the in-modality value for ultrasound. It stays at $83\%$ for CT and $73\%$ for MRI. \revisionfinal{We did not evaluate refinement for the CT holdout, so Table~\ref{tab:transfer} reports only the MRI and US holdouts.} Appendix~\ref{sec:supp_modality} explains the low ultrasound value.}

\rev{Neither experiment shows zero-shot generalisation. We present transfer to unseen anatomies and modalities as an open research direction. An important question is the minimum number of support cases.}

\subsection{Retrieval analysis}
\label{sec:retanalysis}
We first quantify the contribution of the retrieval strategy to refinement performance and then evaluate the retrieved neighbours independently of the refiner.

\paragraph{Retrieval strategy.}
Table~\ref{tab:retrieval_ablation} evaluates the retrieval strategy with the trained \gracem{} refiner held fixed. \revisionfinal{The corrupted input without refinement is the baseline;} a random same-class peer already recovers roughly $40\%$ of the oracle gap, indicating that the anatomically conditioned prior is useful even before any similarity ranking. Selecting the shape-aware K-NN adds a further lift, and the oracle row upper-bounds the retrieval contribution.
%
The gap between the random and K-NN rows measures the combined benefit of similarity-based ranking and the projected retrieval space over random same-class selection; it does not isolate the projector. The projector-versus-raw comparison in Table~\ref{tab:retrieval_projection} isolates that contribution by holding the candidate set and ranking protocol fixed.

\input{tables/retrieval_ablation.tex}

\begin{figure}[!t]
    \centering
    \includegraphics[width=0.98\linewidth]{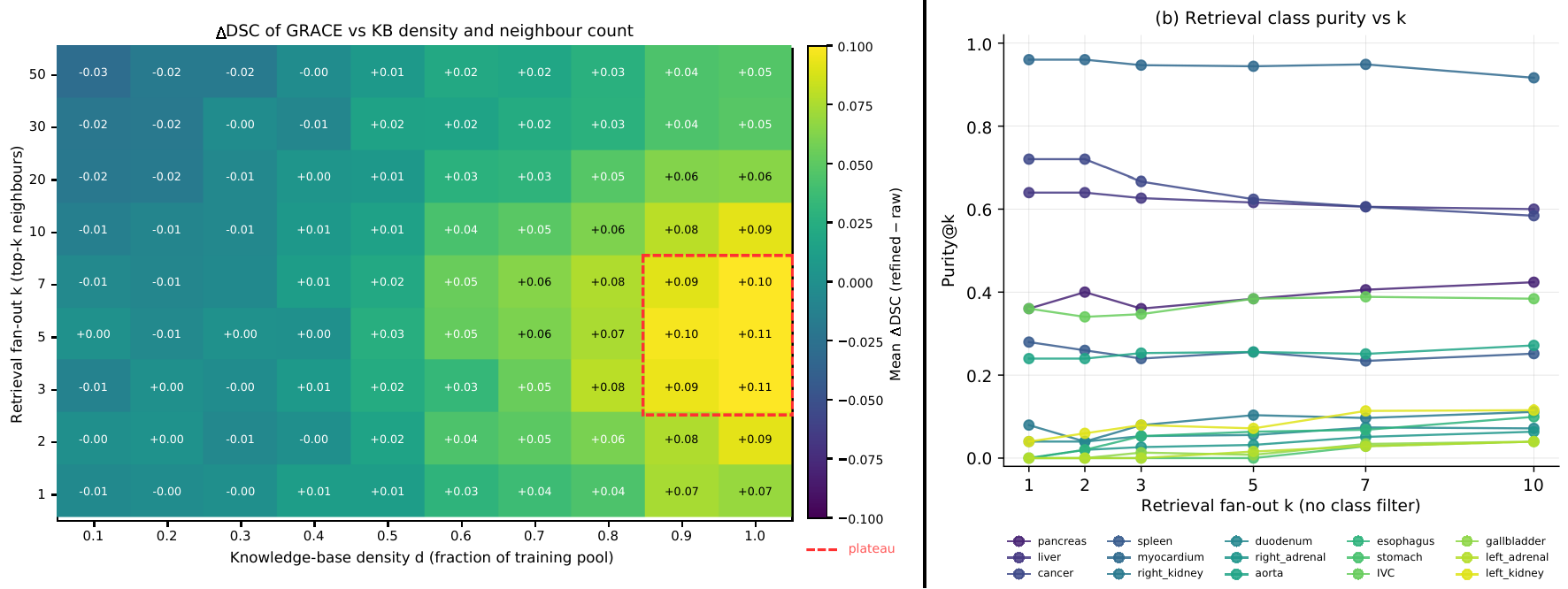}
    \caption{\textbf{Retrieval-module quality.} \emph{(\revisionfinal{a,~}left)} Downstream $\Delta$DSC when varying knowledge-base density ($d$) and retrieval fan-out ($k$). \emph{(\revisionfinal{b,~}right)} Diagnostic class purity@$k$ for the 15 most frequent classes when the usual same-class filter is disabled, one curve per class.}
    \label{fig:retrieval_quality}
\end{figure}

\paragraph{Retrieval-module quality.} 
Figure~\ref{fig:retrieval_quality} characterises two aspects of retrieval: how the knowledge-base density and fan-out affect downstream refinement, and how the unfiltered embedding space groups anatomical classes.
Figure~\ref{fig:retrieval_quality}~(left) reports the $\Delta$DSC improvement obtained for different retrieval configurations, varying the knowledge-base density ($d$) and the retrieval fan-out ($k$). At high density, performance forms a plateau around $k{=}3$--$7$; the main experiments use $k{=}5$ as a point on this empirical plateau.
%
Figure~\ref{fig:retrieval_quality}~(right) deliberately disables the same-class filter and reports class purity@$k$ as an embedding diagnostic. \rev{For a query of class $c$, purity@$k$ is the fraction of its $k$ nearest neighbours in the unfiltered index that also have class $c$. We average this fraction over all queries of class $c$: $\mathrm{purity@}k(c) = \frac{1}{|Q_c|}\sum_{q \in Q_c} \frac{1}{k} \left| \{ j \in \mathrm{NN}_k(q) : \mathrm{class}_j = c \} \right|$. A value of $1$ means every retrieved neighbour already has the correct anatomy without any class filtering.} It does not represent the main candidate pool, for which the per-class indexes of Sec.~\ref{sec:kb} enforce same-class retrieval by construction. The retrieval-encoder ablation is the exception to the main $k{=}5$ setting and uses $k{=}3$, as stated in Table~\ref{tab:embedder_ext} (Appendix~\ref{sec:supp_retrieval}).

\paragraph{Contribution of the exemplar.} We set the retrieval channel of the trained model to zero and keep all other settings. This isolates the part of the gain that the exemplar gives. At the evaluation-only \emph{extreme} severity, \revisionfinal{averaged over all structures,} the exemplar gives $40\%$ of the improvement, and the refiner alone gives the rest. \revisionfinal{On the four organs of Table~\ref{tab:class_query_organs},} the share is from $39\%$ (pancreas) to $62\%$ (kidneys, spleen). In Appendix~\ref{sec:supp_retrieval}, a mask-only encoder is close to MedGemma. Thus, the contribution comes mainly from the exemplar and less from the multimodal retrieval space. We therefore describe \method{} as exemplar-conditioned refinement. The refiner uses the retrieved exemplar as a shape prior, and retrieval selects it from the 15 indexed datasets.

\paragraph{Robustness to a wrong class query.} In a fully automatic pipeline, the class label can be wrong. We keep the trained refiner and compare three query conditions on four organs at four severities. The conditions are the correct class, a random wrong class and no exemplar \revisionfinal{(Table~\ref{tab:class_query_organs})}. At the \emph{extreme} preset, the input DSC is from $0.11$ to $0.31$. The correct class gives the largest gain on every organ. A wrong class gives $0.077$ to $0.151$ DSC less. On the compact organs (liver, kidneys, spleen), a wrong-class exemplar still recovers $50$ to $72\%$ of the correct-class gain over no exemplar. On the pancreas, which is small and variable, a wrong-class exemplar is slightly worse than no exemplar. Thus, the robustness to a wrong class depends on the shape of the target. The stabilisation of refinement under extreme class noise is an open question.

\subsection{Ablation studies}
\label{sec:ablations}
We quantify the contribution of the main design choices underlying \method{} through a series of controlled ablation studies. Specifically, we evaluate the shape-aware retrieval projector while keeping the remainder of the framework unchanged; the choice of retrieval encoder is reported in Appendix~\ref{sec:supp_retrieval} and the comparison of refinement paradigms in Appendix~\ref{sec:refiner_paradigms}.

\paragraph{Shape-aware retrieval projector.} Table~\ref{tab:retrieval_projection} evaluates the contribution of the projector $\phi_\psi$ of Sec.~\ref{sec:proj} across two embedding sources: the MedGemma multimodal embedding used throughout the experiments, and a mask-domain baseline obtained from the frozen VAE with
adaptive pooling and PCA reduction (256).
Within each checkpoint column, the refiner and retrieval protocol are fixed; paired rows for a given encoder source differ only in the similarity space used to rank candidates. On both the full-model and the base checkpoint, and for both encoder sources, the projector yields a comparable positive change in DSC over unprojected cosine (see Table~\ref{tab:retrieval_projection}). The benefit of shape-aware retrieval is therefore largely independent of the underlying encoder.
On unseen query-candidate pairs from the validation split, the Pearson correlation between projected cosine similarity and true shape-Dice increases from $0.42$ (raw MedGemma) to $0.71$ after projection, confirming that the learned space reflects better anatomical similarity.
\input{tables/retrieval_projection.tex}

\subsection{Qualitative overview}
\label{sec:qualitative}
\begin{figure}[!t]
    \centering
    \includegraphics[width=\linewidth]{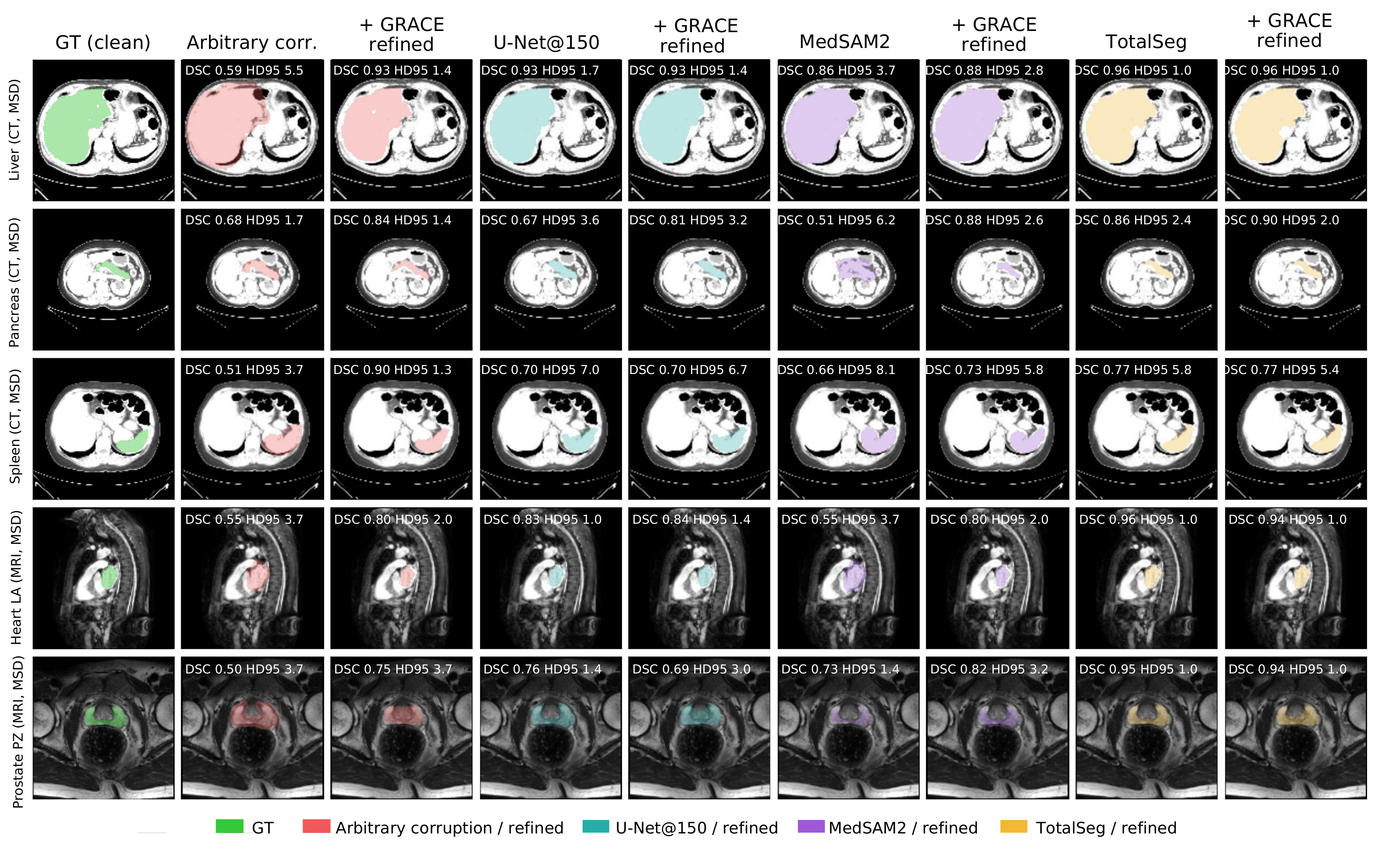}
    \caption{\textbf{Qualitative overview of \method{} refinements.} Rows correspond to Liver, Pancreas, Spleen, Heart LA, and Prostate PZ. Columns show the ground truth; one representative synthetic corruption from the training schedule and its \graceu{} refinement; and three upstream segmentors (U-Net@150, MedSAM2, and TotalSegmentator, with SegResNet used where TotalSegmentator is unavailable) together with their corresponding \graceu{} refinements. Colours identify the source of each mask. DSC and HD95 with respect to the ground truth are overlaid on every panel.}
\label{fig:corruption_examples}
\end{figure}

Figure~\ref{fig:corruption_examples} illustrates representative refinements produced by \method{} across CT (Liver, Pancreas, Spleen) and MRI (Heart LA, Prostate PZ). For each case, one synthetic corruption and three real upstream segmentors (U-Net@150, MedSAM2, and TotalSegmentator or SegResNet) are shown together with the corresponding \graceu{} refinement, with DSC and HD95 overlaid. Three visual patterns recur across the five rows: on the weakest inputs, retrieval provides a strong anatomical prior and recovers organ topology even when the raw prediction contains no usable structure; on intermediate inputs, refinement sharpens boundaries and closes local defects; on strong inputs, refinement makes only minimal changes, consistent with the small deltas of Sec.~\ref{sec:universal}.

\section{Discussion and Conclusions}
\label{sec:discussion}
\paragraph{Discussion.} A central question motivating this work is whether retrieval-conditioned refinement can generalise beyond anatomy-specific correction models. Our experiments indicate that the greatest benefit of retrieval arises when the upstream segmentation is unreliable, precisely the regime in which manual correction is most costly in clinical practice~\citep{galati2022accuracy}, while already accurate predictions receive only minimal changes. \rev{The retrieved exemplar gives $40\%$ of the gain at the most severe preset. The correct class already gives a strong shape prior, and the shape-aware projector adds a smaller gain in that class.}


{\paragraph{When is refinement preferable to a strong segmentor?} We see four settings. (i) No strong segmentor exists for the target. TotalSegmentator covers CT anatomy, but not cardiac chambers in ultrasound or most lesion classes. (ii) The upstream is a black box, for example a vendor model without access to weights or training data. \method{} needs only the output mask. (iii) A new anatomy must be supported without retraining (Sec.~\ref{sec:leave_one_out}). (iv) Promptable models receive imperfect prompts. Our MedSAM2 and ScribblePrompt results use an oracle bounding box. With real prompts, their masks are probably worse and closer to the range where refinement helps. For this reason, we put them in the mid tier. For strong CT segmentors, we do not claim an improvement. \method{} keeps good masks almost unchanged on average, but it degrades some of them (Appendix~\ref{sec:failure}).}

\paragraph{Limitations.} The effectiveness of retrieval depends on the diversity of the knowledge base: anatomies or pathological presentations that are not represented in $\mathcal{K}$ provide weaker retrieval context, and the refiner then remains closer to the raw upstream prediction. Verified support cases can be added to $\mathcal{K}$ without updating the refiner, but genuinely novel morphologies may still require retraining. \rev{At the high end of input quality, $3.8\%$ of the cases with an input DSC above 0.8 lose more than 0.05 DSC (Appendix~\ref{sec:failure}). A mechanism that skips refinement for a probably correct input is a logical next step. \gracem-D{} needs about seven times more compute than \gracem{}, and its results change with the ODE seed. Transfer to a held-out modality is partial on MRI and fails on ultrasound. The leave-one-anatomy-out setting needs $118$ to $302$ verified masks of the held-out organ. We present both as open research directions.}

\paragraph{Conclusions.} We introduced \method{}, a retrieval-conditioned framework that replaces anatomy-specific correctors with a single refiner guided by an anatomy-aware knowledge base. Across upstream segmentors and held-out datasets, it improved unreliable predictions while leaving accurate ones largely unchanged, and verified support cases extend the knowledge base without a parameter update.

\bibliography{main}
\bibliographystyle{tmlr}

\clearpage
\appendix
\section*{Appendix Overview}
Appendix~\ref{sec:refiner_paradigms} compares the three refiner paradigms (deterministic pixel-space, deterministic latent autoencoder, and latent flow-matching) as instantiations of the refiner interface. It reports the mechanistic reasoning behind the empirical ordering in Sec.~\ref{sec:universal}, as well as the associated computational costs of each variant.
Appendix~\ref{sec:arch_ablation} reports the architectural ablation of the latent RAG refiner behind the design choices in Sec.~\ref{sec:refiner_interface}.
\rev{Appendix~\ref{sec:supp_modality} explains the ultrasound result of the leave-one-modality-out experiment. Appendix~\ref{sec:failure} gives the case-level failure analysis. Appendix~\ref{sec:supp_retrieval} gives the retrieval-encoder comparison} \revisionfinal{and the per-organ class-query results}. \revisionfinal{Appendix~\ref{sec:supp_datasets} summarises the dataset scope, per-dataset case and class counts, and the aggregate number of non-empty binary-mask instances.} Finally, Appendix~\ref{sec:supp_longitudinal} reports an emergent temporal-memory property of the retrieval pipeline on the LUMIERE dataset.

\section{Refiner paradigms: deterministic, latent autoencoder, latent flow-matching}
\label{sec:refiner_paradigms}
The refiner interface (Sec.~\ref{sec:refiner_interface}) admits three concrete instantiations. They are evaluated with the same data, corruption schedule, and retrieval pipeline, but differ in network architecture, refinement representation (pixel or latent), training objective, and inference procedure (deterministic forward pass or generative sampling).

The first instantiation, \graceu{}, is a supervised 3D U-Net trained on pixel-wise labels that consumes the corrupted mask, the image, and the retrieved exemplar as channel-concatenated inputs and returns a per-voxel foreground logit. It uses one forward pass with no latent codec or sampler. Its single-pass design makes it the fastest variant and empirically produces the more conservative behaviour described in Sec.~\ref{sec:universal}. Consistent with Table~\ref{tab:universal}, \gracem{} produces the larger mean DSC gains on the \revisionfinal{mid tier and on two of the three strong segmentors, whereas \graceu{} leads on the weak tier and on U-Net@150; all three variants} have small non-negative mean changes on the strong tier.

The second instantiation, \gracem{}, moves the refinement into the mask latent space. A frozen encoder maps the corrupted mask and the retrieved exemplar into a compact $4{\times}\!24^3$ latent. A small fusion module (concat, FiLM, or cross-attention) combines the two latents, while a mask-tuned decoder returns the refined mask. Since the decoded output is drawn from the learned mask manifold, this paradigm can produce a plausible completion even when the input carries almost no signal. The reported latent variants use a frozen MAISI encoder together with a decoder fine-tuned on the binary-mask distribution (Appendix~\ref{sec:arch_ablation}).


The third instantiation, \gracem-D{}, replaces the deterministic latent decode with an ODE integration of a learned velocity field conditioned on the retrieved exemplar, following \revisionfinal{flow matching~\citep{lipman2022flow}} for latent generative models. Inference becomes stochastic in the choice of initial noise, so a batch of runs on the same input yields a family of plausible refinements; averaging or selecting among them supports the sample-level uncertainty estimates that neither deterministic paradigm provides. On the aggregate benchmark in Table~\ref{tab:det_vs_gen}, its 30-step aggregate DSC closely matches the deterministic \gracem{} decode while inference is roughly seven times slower; its practical value therefore lies in downstream workflows that consume uncertainty rather than in additional aggregate DSC.

These architectural differences are reflected in the empirical behaviour observed across experiments.
Table~\ref{tab:universal} supports regime-wise comparisons for \rev{the three variants}: \gracem{} has the larger mean lift on the mid and most of the strong tiers, while \revisionfinal{all three variants} have small non-negative mean changes on the top performing model (TotalSegmentator). \revisionfinal{On the aggregate benchmark of Table~\ref{tab:det_vs_gen}, averaged over three ODE seeds, the 30-step \gracem-D{} stays within $0.01$ DSC of \gracem{}.}

\paragraph{Inference runtime and computational cost.} Table~\ref{tab:det_vs_gen} reports wall-clock inference time per case and mean DSC for the three refiner paradigms. Measurements were obtained on a single NVIDIA A100 with batch size one using ten representative cases (one per MSD organ) from a stratified 10\% hold-out. Each benchmark run covers those ten cases; runtime means and sample standard deviations are computed across five repeated runs, with GPU synchronisation before each measurement.
\graceu{} runs sub-second per case using one forward pass, and \gracem{} adds a single deterministic latent decode. \gracem-D{} requires several seconds at the 30-step operating point because of iterative ODE integration (see Table~\ref{tab:det_vs_gen}). All three variants share the same retrieval pipeline, allowing deployments to switch between refiners without modifying the surrounding retrieval infrastructure.

\begin{table}[t]
  \centering
  \caption{\textbf{Performance versus inference runtime.} Wall-clock time per case (mean $\pm$ sample standard deviation across five benchmark runs, each covering ten organs) and DSC (mean $\pm$ sample standard deviation across cases) under the strong synthetic-corruption preset. PostDAE requires 10 separately trained models (one per anatomy). The number of function evaluations (NFE) is also reported.
  }
  \label{tab:det_vs_gen}
  \setlength{\tabcolsep}{7pt}
  \small
  \begin{tabular}{lccc}
    \toprule
    Method & NFE & Time / case (s) & DSC $\uparrow$ \\
    \midrule
    \graceu{} (feedforward)          & 1  & $0.31 \pm 0.02$ & $0.803 \pm 0.071$ \\
    \gracem{} (deterministic latent) & 1  & $0.58 \pm 0.03$ & $0.815 \pm 0.068$ \\
    \gracem-D{} (FM, 10 steps)       & 10 & $1.4 \pm 0.1$   & $0.788 \pm 0.073$ \\
    \gracem-D{} (FM, 20 steps)       & 20 & $2.8 \pm 0.1$   & $0.801 \pm 0.070$ \\
    \gracem-D{} (FM, 30 steps)       & 30 & $4.2 \pm 0.1$   & $0.810 \pm 0.069$ \\
    \gracem-D{} (FM, 50 steps)       & 50 & $7.0 \pm 0.2$   & $0.813 \pm 0.068$ \\
    \midrule
    PostDAE$^{\dagger}$ & 1 & $0.16 \pm 0.11$ & $0.422 \pm 0.132$ \\
    RAM$^{\ddagger}$     & 1 & $10.7 \pm 0.1$  & $0.582 \pm 0.108$ \\
    \bottomrule
    \multicolumn{4}{l}{$^{\dagger}$Slice-by-slice 2-D processing; 10 anatomy-specific models.}\\
    \multicolumn{4}{l}{$^{\ddagger}$Slice-by-slice 2-D processing; single shared model.}\\
  \end{tabular}
\end{table}



\section{Architectural ablations of the latent RAG refiner}
\label{sec:arch_ablation}
The refiner interface of Sec.~\ref{sec:refiner_interface} admits several concrete instantiations. This appendix reports the empirical characterisation that motivates the choices adopted in the main paper. Three design axes are examined in controlled comparisons: (i) how the retrieved exemplar $Y_r$ enters the refiner (\emph{fusion module}); (ii) whether inference uses the deterministic \gracem{} decode or the \gracem-D{} flow-matching trajectory; and (iii) the training loss, exemplar selection, and velocity-network width for \gracem-D{}. Settings outside the axis under study are held fixed within each comparison. The same fusion-module ablation is repeated on the pixel-space variant \graceu{} for completeness. 

\paragraph{Fusion module for RAG.} Three interfaces for injecting the retrieved exemplar into the refiner are compared. The \emph{concat} module stacks the encoded corrupted latent and the encoded exemplar latent along the channel axis and applies a $1{\times}1{\times}1$ convolution with identity initialisation so the refiner recovers the corrupted-only forward pass at step zero. The \emph{FiLM} module pools the encoded exemplar over the spatial dimensions, passes it through a small MLP that emits per-channel scale and shift, and applies them to the corrupted latent. The \emph{cross-attention} module is a single-head projected attention layer with the corrupted latent tokens as queries and the encoded exemplar tokens as keys and values. The pixel-space ablation applies the corresponding conditioning operations to U-Net feature maps. Table~\ref{tab:fusion_module} reports refined DSC for both refiner variants. Concat gives the highest score for each variant and is retained as the default for its accuracy, parameter economy, and interpretability. Additionally, Table~\ref{tab:loss_ablation} reports controlled combinations of training changes for \gracem-D{}: the auxiliary Dice loss and the K-NN exemplar each improve the common baseline, their combination improves it further, and widening the velocity network produces the highest score at roughly twice the parameter count.

\begin{table}[t]
  \centering
  \caption{\textbf{Fusion-module ablation.} Refined DSC (mean $\pm$ sample standard deviation across cases) for the three exemplar-fusion
  interfaces, evaluated on both refiner variants under the strong synthetic-corruption preset. All rows share
  the training data, corruption schedule, and retrieval pipeline; only the module that
  injects the retrieved exemplar changes. Concat gives the highest displayed score for both variants and is retained as the default.}
  \label{tab:fusion_module}
  \small
  \setlength{\tabcolsep}{10pt}
  \begin{tabular}{lcc}
    \toprule
    Fusion module & \graceu{} DSC $\uparrow$ & \gracem{} DSC $\uparrow$ \\
    \midrule
    Concat (default)  & $0.803 \pm 0.071$ & $0.815 \pm 0.068$ \\
    FiLM              & $0.786 \pm 0.074$ & $0.797 \pm 0.072$ \\
    Cross-attention   & $0.750 \pm 0.086$ & $0.799 \pm 0.071$ \\
    \bottomrule
  \end{tabular}
\end{table}

\input{tables/loss_ablation.tex}

\paragraph{Decoder handling.} All reported latent variants use the frozen MAISI encoder with a decoder fine-tuned on the binary-mask distribution; no frozen-decoder comparison is included in the reported ablations.

\paragraph{Inference procedure.} On the aggregate benchmark in Table~\ref{tab:det_vs_gen}, the 30-step flow-matching procedure reaches $0.810$ DSC versus $0.815$ for the deterministic \gracem{} latent decode and is about seven times slower. The deterministic procedure is therefore used for the reported \gracem{} numbers.

\section{Leave-one-modality-out: retrieval on ultrasound}
\label{sec:supp_modality}
\rev{This appendix explains the ultrasound result of Table~\ref{tab:transfer} (Block B). Class routing does not cause the ultrasound failure: all 101 CAMUS queries retrieve the correct cardiac structure. The shape agreement between a query and its retrieved pool decreases. The low ultrasound ratio has a specific cause. CAMUS is 2D+time, so its third axis is the cardiac cycle. ACDC and MSD Heart are spatial volumes. The CAMUS long-axis views (2CH, 4CH) also retrieve short-axis support cases. The shape agreement per structure follows the view dependence of the structure: myocardium $0.125$, left atrium $0.239$, LV cavity $0.433$. We think that shape complexity is an important factor in retrieval. Thus, the ultrasound row does not measure the modality alone.}

\section{Case-level failure analysis}
\label{sec:failure}
\rev{A mean change does not show how often refinement degrades a mask. Table~\ref{tab:universal} gives the mean $\Delta$DSC and its standard deviation for each upstream tier. Table~\ref{tab:harm_bands} groups the cases by input DSC and gives how often \method{} decreases the DSC.} \revisionfinal{Over all $800$ cases, $5.5\%$ ($44$) lose more than $0.01$ DSC and $1.5\%$ ($12$) lose more than $0.05$. The 5th percentile of $\Delta$DSC therefore lies between $-0.05$ and $-0.01$. For inputs below $0.5$ DSC, fewer than $5\%$ of the cases lose more than $0.01$, so the 5th percentile is above $-0.01$.}

\input{tables/harm_bands.tex}

\rev{Below an input DSC of $0.8$, harm is rare. At most $6.6\%$ of cases lose more than $0.01$ DSC, and at most $1.2\%$ lose more than $0.05$. Above $0.8$, the rate increases to $11.5\%$, and $3.8\%$ of cases lose more than $0.05$. A good input has few errors to correct, and the shape prior can remove correct detail. On strong upstreams, the harm rate is more informative than the mean.}

\section{Retrieval encoder choice}
\label{sec:supp_retrieval}

\paragraph{Retrieval encoder.} 
Table~\ref{tab:embedder_ext} evaluates the impact of the retrieval encoder while keeping the refiner and the knowledge-base size fixed. We compare two families of retrieval embeddings: four pretrained foundation encoders (DINOv2~\citep{oquab2023dinov2}, BiomedCLIP~\citep{zhang2023biomedclip}, MedSigLIP~\citep{sellergren2025medgemma}, and MedGemma) and three mask-domain variants derived from a frozen MAISI~\citep{guo2025maisi} VAE encoder. The three mask-domain variants use (i) pooled latent features directly, (ii) pooled features followed by PCA reduction, and (iii) a learned two-layer self-projector trained on the same latent representation. Among foundation encoders, MedGemma achieves the best performance, followed closely by the remaining medical encoders, whereas DINOv2 consistently trails them. The strongest mask-domain variant closely approaches MedGemma despite relying solely on anatomical shape information, suggesting that shape alone carries a strong retrieval signal.

\input{tables/embedder_ablation.tex}

\paragraph{\revisionfinal{Class query.}} \revisionfinal{Table~\ref{tab:class_query_organs} gives the per-organ results of the class-query study of Sec.~\ref{sec:retanalysis} at the \emph{extreme} preset.}

\begin{table}[t]
  \centering
  \caption{\revisionfinal{\textbf{Robustness to the class query.} Mean DSC gain over the input at the evaluation-only \emph{extreme} preset, with the trained refiner held fixed. \emph{Correct class} is the protocol used everywhere else in the paper. \emph{Random wrong class} retrieves the exemplar from a randomly chosen other class. \emph{No exemplar} sets the retrieval channel to zero. The exemplar share is (correct class $-$ no exemplar) / correct class.}}
  \label{tab:class_query_organs}
  \small
  \setlength{\tabcolsep}{7pt}
  \revisionfinal{\begin{tabular}{lccccc}
    \toprule
    Organ & Input DSC & Correct class & Random wrong class & No exemplar & Exemplar share \\
    \midrule
    Liver    & $0.183$ & $+0.512$ & $+0.361$ & $+0.208$ & $59\%$ \\
    Kidneys  & $0.110$ & $+0.449$ & $+0.372$ & $+0.171$ & $62\%$ \\
    Spleen   & $0.201$ & $+0.430$ & $+0.338$ & $+0.163$ & $62\%$ \\
    Pancreas & $0.308$ & $+0.194$ & $+0.101$ & $+0.119$ & $39\%$ \\
    \bottomrule
  \end{tabular}}
\end{table}

\section{Datasets and instance counts}
\label{sec:supp_datasets}
Table~\ref{tab:dataset_scope} enumerates the full dataset scope, listing per-dataset
modality, case count, class count, and the anatomical structures covered for both the
training corpus and the three held-out datasets. Decomposing the multi-class training
labels into per-organ binaries and discarding empty class--volume pairs yields \revisionfinal{11,900} non-empty 3-D binary-mask instances
over 38 structures \revisionfinal{in total, of which the knowledge base of the reported configuration indexes $N{=}3{,}913$}. As stated in Sec.~\ref{sec:kb}, the
held-out-dataset evaluation indexes only training-split cases; the optional
longitudinal-memory analysis in Appendix~\ref{sec:supp_longitudinal} is reported
separately.

\input{tables/dataset_scope.tex}


\section{Retrieval as implicit temporal memory}
\label{sec:supp_longitudinal}
\begin{center}
    \centering
    \includegraphics[width=\linewidth]{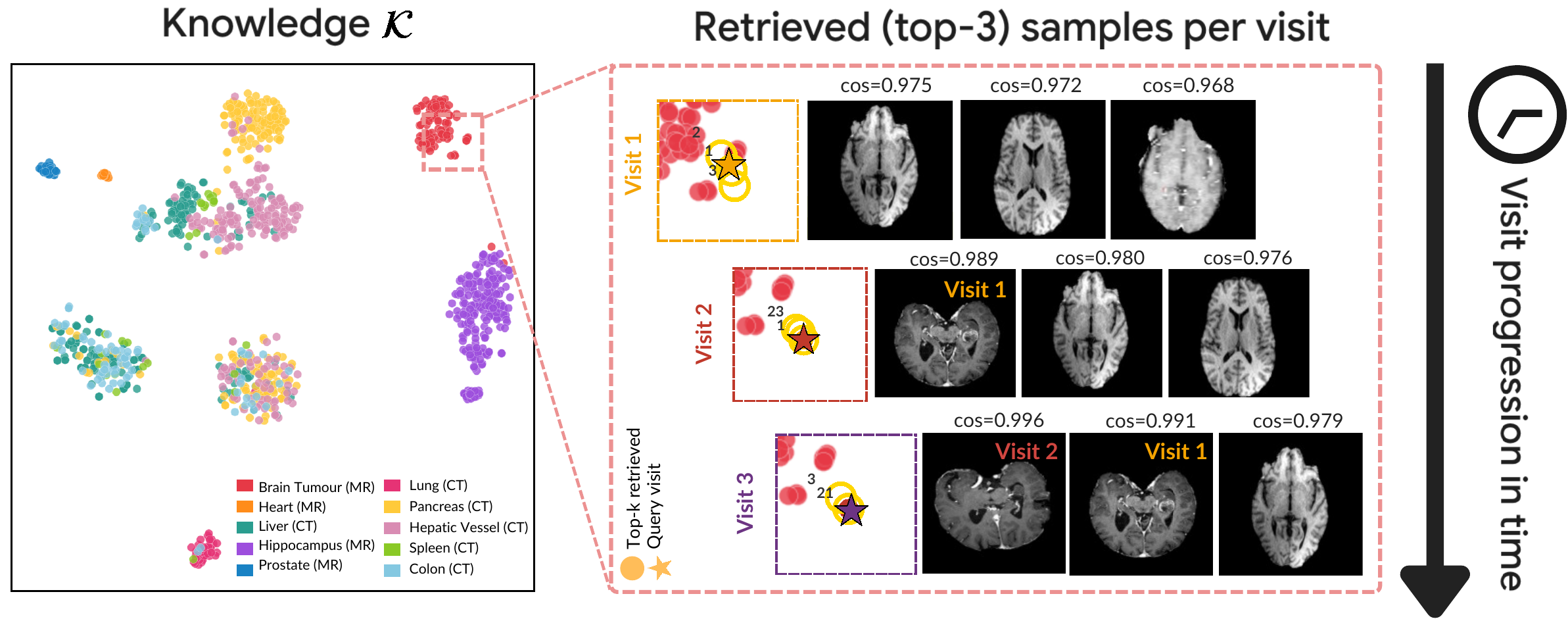}
    \captionsetup{hypcap=false}
    \captionof{figure}{\textbf{Emergent patient-specific retrieval (LUMIERE).} Each row corresponds to one visit as prior scans are added to $\mathcal{K}$. \emph{Left:} UMAP of training-pool embeddings with three patient visits ($\star$). \emph{Right:} top-3 retrieved neighbours (gold rings) and similarity scores.}
    \label{fig:three_visit}
\end{center}

An emergent behaviour of the retrieval pipeline appears on the longitudinal LUMIERE dataset when the temporal-memory workflow is enabled: as a patient's visits are embedded into $\mathcal{K}$, later refinements retrieve the patient's own prior scans instead of cross-patient atlases (Fig.~\ref{fig:three_visit}). At the first visit, retrieved neighbours are tumour examples from other patients in the training pool; once the patient's first scan is indexed, it becomes the highest-ranked retrieval at the next visit; by the third visit, prior visits dominate the top-$k$ list. No temporal supervision is used: retrieval is driven by similarity in the embedding space. This optional workflow is separate from Table~\ref{tab:unseen}, whose held-out-dataset protocol uses the fixed training-only index and therefore excludes every LUMIERE case from the candidate pool. On the subset of LUMIERE with at least one prior visit ($n{=}508$ timepoints), enabling same-patient retrieval yields an additional DSC lift of $+0.021$ on the ScribblePrompt upstream\revisionfinal{, on top of the fixed-index score of the corresponding row of Table~\ref{tab:unseen}}; whether this lift is admissible depends on the clinical workflow (a longitudinal QC pass at visit $t$ that already trusts the visit-$(t{-}1)$ mask, versus a per-visit review pipeline that does not) and is reported separately rather than folded into the headline number. The mechanism acts as an implicit temporal memory and supports longitudinal workflows without dedicated temporal architectures.


\end{document}

%% file: math_commands.tex
\usepackage{amsmath,amsfonts,bm}

\def\eqref#1{equation~\ref{#1}}

\def\1{\bm{1}}

\DeclareMathAlphabet{\mathsfit}{\encodingdefault}{\sfdefault}{m}{sl}
\SetMathAlphabet{\mathsfit}{bold}{\encodingdefault}{\sfdefault}{bx}{n}



%% file: tables/universal.tex
\begin{table}[!t]
\centering
\caption{
\method{} as a universal refiner across upstream operating regimes. Weak, mid, and strong upstream segmentors are evaluated using DSC ($\uparrow$) and HD95 ($\downarrow$). Raw upstream rows report absolute scores. Refinement methods report signed changes ($\Delta$) relative to the corresponding upstream prediction (mean $\pm$ standard deviation). $^\dagger$ denotes statistical significance against the raw upstream (paired Wilcoxon, $p<0.05$). Bold indicates the best refinement change per column; ``--'' denotes configurations that are not applicable. \rev{The latent flow-matching variant \gracem-D{} is reported alongside the two deterministic variants so that all three instantiations can be compared directly; its inference cost is \revisionfinal{about seven times} higher (Table~\ref{tab:det_vs_gen}).}}
%
\label{tab:universal}
\scriptsize
\setlength{\tabcolsep}{9pt}
\renewcommand{\arraystretch}{1.1}
\begin{tabular}{l c c c c c}
    \toprule
    & Weak & Mid & \multicolumn{3}{c}{Strong} \\
    \cmidrule(lr){2-2}\cmidrule(lr){3-3}\cmidrule(lr){4-6}
    Refiner & U-Net@30 & MedSAM2 & TotalSeg & SegResNet & U-Net@150 \\
    \midrule
    \multicolumn{6}{l}{\emph{Block A --- DSC ($\uparrow$)}} \\
    \midrule
    Raw upstream & $0.386$ & $0.727$ & $0.932$ & $0.793$ & $0.775$ \\
    \midrule
    Morphological
      & $+0.005 \pm 0.019$ & $+0.003 \pm 0.021$
      & $-0.005 \pm 0.011$ & $-0.002 \pm 0.014$ & $-0.003 \pm 0.015$ \\
    TTAug
      & $+0.018 \pm 0.024$ & ---
      & $+0.002 \pm 0.008$ & $+0.007 \pm 0.013$ & $+0.008 \pm 0.014$ \\
    PostDAE
      & $-0.108 \pm 0.117$ & $-0.330 \pm 0.140$
      & $-0.487 \pm 0.152$ & $-0.294 \pm 0.148$ & $-0.221 \pm 0.144$ \\
    RAM
      & $+0.002 \pm 0.003$ & $+0.002 \pm 0.002$
      & $+0.001 \pm 0.003$ & $+0.002 \pm 0.002$ & $+0.001 \pm 0.003$ \\
    U-Net w/o RAG
      & $+0.049 \pm 0.076$ & $+0.017 \pm 0.038$
      & $-0.006 \pm 0.019$ & $+0.009 \pm 0.022$ & $+0.011 \pm 0.024$ \\
    \graceu{}
      & $\mathbf{+0.353^{\dagger} \pm 0.091}$ & $+0.061^{\dagger} \pm 0.058$
      & $\phantom{+}0.000 \pm 0.017$ & $+0.014 \pm 0.023$ & $\mathbf{+0.021 \pm 0.019}$ \\
    \gracem{}
      & $+0.342^{\dagger} \pm 0.096$ & $\mathbf{+0.083^{\dagger} \pm 0.052}$
      & $\mathbf{+0.008 \pm 0.016}$ & $\mathbf{+0.024^{\dagger} \pm 0.021}$ & $+0.019^{\dagger} \pm 0.020$ \\
    \rev{\gracem-D{}}
      & \rev{$+0.336^{\dagger} \pm 0.098$} & \rev{$+0.078^{\dagger} \pm 0.054$}
      & \rev{$+0.006 \pm 0.017$} & \rev{$+0.021^{\dagger} \pm 0.022$} & \rev{$+0.017^{\dagger} \pm 0.021$} \\
    \midrule
    \multicolumn{6}{l}{\emph{Block B --- HD95 (mm, $\downarrow$)}} \\
    \midrule
    Raw upstream & $21.3$ & $10.4$ & $\phantom{0}4.1$ & $\phantom{0}8.2$ & $\phantom{0}9.4$ \\
    \midrule
    Morphological
      & $-0.4 \pm 1.6$ & $-0.2 \pm 1.1$
      & $+0.1 \pm 0.6$ & $\phantom{+}0.0 \pm 0.9$ & $\phantom{+}0.0 \pm 1.0$ \\
    TTAug
      & $-0.7 \pm 1.9$ & ---
      & $-0.1 \pm 0.5$ & $-0.2 \pm 0.8$ & $-0.3 \pm 0.9$ \\
    PostDAE
      & $+4.2 \pm 4.6$ & $+6.7 \pm 5.4$
      & $+10.2 \pm 6.1$ & $+6.4 \pm 5.6$ & $+5.5 \pm 5.4$ \\
    RAM
      & $\phantom{+}0.0 \pm 0.2$ & $\phantom{+}0.0 \pm 0.2$
      & $\phantom{+}0.0 \pm 0.1$ & $\phantom{+}0.0 \pm 0.1$ & $\phantom{+}0.0 \pm 0.1$ \\
    U-Net w/o RAG
      & $+1.6 \pm 3.0$ & $+2.4 \pm 2.9$
      & $+2.9 \pm 2.5$ & $+2.1 \pm 2.4$ & $+1.9 \pm 2.4$ \\
    \graceu{}
      & $\mathbf{-10.4^{\dagger} \pm 5.0}$ & $-2.7^{\dagger} \pm 2.5$
      & $-0.3 \pm 0.8$ & $-0.9^{\dagger} \pm 1.4$ & $-0.9^{\dagger} \pm 1.5$ \\
    \gracem{}
      & $-9.8^{\dagger} \pm 5.1$ & $\mathbf{-3.4^{\dagger} \pm 2.6}$
      & $\mathbf{-0.5^{\dagger} \pm 0.7}$ & $\mathbf{-1.1^{\dagger} \pm 1.5}$ & $\mathbf{-1.0^{\dagger} \pm 1.5}$ \\
    \rev{\gracem-D{}}
      & \rev{$-9.5^{\dagger} \pm 5.2$} & \rev{$-3.1^{\dagger} \pm 2.6$}
      & \rev{$-0.4 \pm 0.8$} & \rev{$-1.0^{\dagger} \pm 1.5$} & \rev{$-0.9^{\dagger} \pm 1.5$} \\
    \bottomrule
\end{tabular}
\end{table}

%% file: tables/unseen.tex
\begin{table}[!t]
\centering
%
\caption{\textbf{\revisionfinal{Held-out dataset transfer to three datasets.}} Raw upstream and refined DSC and HD95 (mm) for \revisionfinal{three upstream segmentors (one or two per dataset)}. No case from a held-out dataset is inserted into $\mathcal{K}$; every query retrieves only from the fixed training-corpus index. $\Delta$ denotes refined $-$ raw performance (positive is better for DSC, negative for HD95). Best refiner per row is shown in \textbf{bold}.}
\label{tab:unseen}
\small
\setlength{\tabcolsep}{5pt}
\renewcommand{\arraystretch}{1.15}
\resizebox{\textwidth}{!}{
\begin{tabular}{ll cc cc cc}
\toprule
& & \multicolumn{2}{c}{Raw} & \multicolumn{2}{c}{$+$\graceu{}} & \multicolumn{2}{c}{$+$\gracem{}} \\
\cmidrule(lr){3-4}\cmidrule(lr){5-6}\cmidrule(lr){7-8}
Dataset & Segmentor & DSC & HD95 & $\Delta$DSC & $\Delta$HD95 & $\Delta$DSC & $\Delta$HD95 \\
\midrule
\multirow{1}{*}{ProstateX}
  & ScribblePrompt  & $0.753 \pm 0.078$ & $8.1 \pm 2.6$ & $\boldsymbol{+0.041 \pm 0.034}$ & $\boldsymbol{-2.4 \pm 1.5}$ & $+0.036 \pm 0.036$ & $-2.0 \pm 1.6$ \\
\midrule
\multirow{2}{*}{\revisionfinal{CHAOS}}
  & TotalSeg (fast) & $0.921 \pm 0.041$ & $3.2 \pm 1.1$ & $+0.006 \pm 0.017$ & $-0.4 \pm 0.6$ & $\boldsymbol{+0.009 \pm 0.019}$ & $\boldsymbol{-0.5 \pm 0.7}$ \\
  & SwinUNETR-BTCV & $0.874 \pm 0.052$ & $4.7 \pm 1.7$ & $\boldsymbol{+0.019 \pm 0.024}$ & $\boldsymbol{-1.0 \pm 0.9}$ & $+0.017 \pm 0.025$ & $-0.9 \pm 1.0$ \\
\midrule
\multirow{1}{*}{LUMIERE}
  & ScribblePrompt  & $0.716 \pm 0.084$ & $11.5 \pm 3.4$ & $+0.062 \pm 0.041$ & $-3.6 \pm 1.9$ & $\boldsymbol{+0.079 \pm 0.044}$ & $\boldsymbol{-4.2 \pm 2.1}$ \\
\bottomrule
\end{tabular}}
\end{table}

%% file: tables/transfer.tex
\begin{table}[!t]
\centering
\caption{\rev{\textbf{Transfer to a held-out anatomy or modality.} In both blocks, we remove the held-out anatomy or modality from the training data and from the retrieval index, and we retrain the refiner from scratch. Block A: we add the stated number of verified support cases of the held-out organ to $\mathcal{K}$ and do not update the refiner. The support cases are disjoint from the queries. Block B: no case of the held-out modality enters the index, and one \revisionfinal{\graceu{}} refiner is evaluated per modality. The control is the refiner trained with that modality, on the same queries. Shape ret.\ is the query--pool shape agreement without the modality, relative to the agreement with it. $\Delta$ is refined $-$ input. ``--'' denotes configurations that are not applicable or not evaluated.}}
\label{tab:transfer}
\scriptsize
\setlength{\tabcolsep}{4pt}
\renewcommand{\arraystretch}{1.1}
\begin{tabular}{l l l c c c c c}
    \toprule
    Held-out & Query & Index & Input DSC & \graceu{} $\Delta$DSC & \gracem{} $\Delta$DSC & Control $\Delta$ & Shape ret. \\
    \midrule
    \multicolumn{8}{l}{\emph{Block A --- Leave-one-anatomy-out (parameter-free index extension)}} \\
    \midrule
    Pancreas    & U-Net@30 & $+241$ & $0.013 \pm 0.019$ & $+0.164 \pm 0.078$ & $\mathbf{+0.181 \pm 0.081}$ & -- & -- \\
    Spleen      & U-Net@30 & $+273$ & $0.033 \pm 0.025$ & $+0.238 \pm 0.091$ & $\mathbf{+0.246 \pm 0.088}$ & -- & -- \\
    Liver       & U-Net@30 & $+302$ & $0.130 \pm 0.056$ & $\mathbf{+0.317 \pm 0.094}$ & $+0.301 \pm 0.096$ & -- & -- \\
    Prostate PZ & U-Net@30 & $+118$ & $0.421 \pm 0.113$ & $\mathbf{+0.088 \pm 0.061}$ & $+0.082 \pm 0.064$ & -- & -- \\
    Mean        &          &                & $0.149$ & $+0.202$ & $\mathbf{+0.203}$ & -- & -- \\
    \midrule
    \multicolumn{8}{l}{\emph{Block B --- Leave-one-modality-out}} \\
    \midrule
    MRI & CHAOS MRI liver & 713 CT livers & $0.515$ & \revisionfinal{$+0.084$} & \revisionfinal{--} & $+0.349$ & $73\%$ \\
    US  & CAMUS cardiac   & CT\,+\,MRI    & $0.438$ & \revisionfinal{$-0.202$} & \revisionfinal{--} & $+0.334$ & $29\%$ \\
    \bottomrule
\end{tabular}
\end{table}

%% file: tables/retrieval_ablation.tex
\begin{table}[!t]
\centering
\caption{\textbf{Retrieval-strategy assessment.}
All rows are evaluated on the strong synthetic-corruption preset. \revisionfinal{The first row is the corrupted input without refinement (baseline). The other three rows use} the same trained \gracem{} refiner and differ only
in the same-class mask provided as the spatial exemplar channel. The oracle row
retrieves the ground-truth mask and provides an upper bound on the contribution of retrieval to refinement.}
\label{tab:retrieval_ablation}
\small
\setlength{\tabcolsep}{6pt}
\begin{tabular}{lcc}
\toprule
Retrieval strategy & DSC $\uparrow$ & HD95 (voxels) $\downarrow$ \\
\midrule
\revisionfinal{Corrupted input (no refinement, baseline)} & $0.660 \pm 0.104$ & $10.42 \pm 4.31$ \\
\midrule
Random same-class peer & $0.755 \pm 0.087$ & $8.16 \pm 3.62$ \\
K-NN same-class peer (\textbf{ours}) & $0.815 \pm 0.068$ & $6.71 \pm 2.98$ \\
Oracle (retrieve GT) [upper bound] & $0.881 \pm 0.052$ & $4.94 \pm 2.11$ \\
\bottomrule
\end{tabular}
\end{table}

%% file: tables/retrieval_projection.tex
\begin{table}[!t]
\centering
\caption{\textbf{Shape-aware retrieval projector.}
Performance on the strong synthetic-corruption preset using two embedding sources:
MedGemma multimodal embeddings and a mask-domain baseline derived from the frozen VAE
encoder. \emph{Base} and \emph{full model} denote two fixed \revisionfinal{\gracem-D{}} refiner configurations \revisionfinal{(we report one variant for simplicity)};
within each column, only the encoder source and similarity space change.
\emph{Shape-proj.} denotes the learned shape-aware projection.}
\label{tab:retrieval_projection}
\small
\setlength{\tabcolsep}{5pt}
\resizebox{\textwidth}{!}{
\begin{tabular}{llcccc}
\toprule
Encoder source & Similarity space & Dim & DSC (base) & DSC (full model) & $\Delta_{\mathrm{full}}$ vs.\ raw cosine \\
\midrule
MedGemma-multimodal & unprojected cosine (control) & 2560 & $0.708 \pm 0.082$ & $0.741 \pm 0.076$ & --- \\
MedGemma-multimodal & \textbf{shape-proj. (ours)} & 256 & $0.727 \pm 0.079$ & $0.771 \pm 0.072$ & $+0.030$ \\
\midrule
Mask-domain (VAE + pool + PCA) & unprojected cosine (control) & 256 & $0.718 \pm 0.084$ & $0.752 \pm 0.075$ & --- \\
Mask-domain (VAE + pool + PCA) & \textbf{shape-proj. (ours)} & 256 & $0.739 \pm 0.078$ & $0.782 \pm 0.071$ & $+0.030$ \\
\bottomrule
\end{tabular}
}
\end{table}

%% file: tables/loss_ablation.tex
\begin{table}[!t]
\centering
\caption{\textbf{\revisionfinal{Training ablation} for \gracem-D{}.}
Rows compare controlled combinations against a common baseline on the strong
synthetic-corruption preset; $\Delta$DSC is measured relative to that baseline. Params
report only the U-Net velocity network.}
\label{tab:loss_ablation}
\small
\setlength{\tabcolsep}{6pt}
\begin{tabular}{lccc}
\toprule
Configuration & DSC $\uparrow$ & $\Delta$DSC & Params (M) \\
\midrule
Baseline & $0.706 \pm 0.082$ & $0.000 \pm 0.000$ & \real{53.2} \\
Dice-aux only ($w{=}0.5$, 60-ep warmup) & $0.721 \pm 0.079$ & $+0.015 \pm 0.021$ & \real{53.2} \\
K-NN exemplar only & $0.717 \pm 0.081$ & $+0.011 \pm 0.020$ & \real{53.2} \\
Dice-aux + K-NN exemplar  & $0.732 \pm 0.076$ & $+0.026 \pm 0.024$ & \real{53.2} \\
Both + wider U-Net ($\times 2$ channels) & $0.744 \pm 0.073$ & $+0.038 \pm 0.026$ & 118.6 \\
\bottomrule
\end{tabular}
\end{table}

%% file: tables/harm_bands.tex
\begin{table}[!t]
\centering
\caption{\rev{\textbf{Case-level failure analysis.} Evaluated cases grouped by
the DSC of their input mask. A case is harmed when refinement lowers its DSC by
more than the stated margin; case counts in brackets. \revisionfinal{Mean $\Delta$DSC is the mean change of DSC within each band.}}}
\label{tab:harm_bands}
\small
\setlength{\tabcolsep}{9pt}
\renewcommand{\arraystretch}{1.1}
\begin{tabular}{lcccc}
\toprule
Input DSC & $n$ & \revisionfinal{Mean $\Delta$DSC} & Harmed ($>0.01$ drop) & Harmed ($>0.05$ drop) \\
\midrule
$<0.2$         & $214$ & \revisionfinal{$+0.271$} & $1.4\%$ \,(3)  & $0.5\%$ \,(1) \\
$0.2$ to $0.5$ & $187$ & \revisionfinal{$+0.168$} & $3.7\%$ \,(7)  & $1.1\%$ \,(2) \\
$0.5$ to $0.8$ & $243$ & \revisionfinal{$+0.061$} & $6.6\%$ (16)   & $1.2\%$ \,(3) \\
$>0.8$         & $156$ & \revisionfinal{$+0.007$} & $11.5\%$ (18)  & $3.8\%$ \,(6) \\
\bottomrule
\end{tabular}
\end{table}

%% file: tables/embedder_ablation.tex
\begin{table}[!t]
\centering
\caption{
\textbf{Retrieval encoder ablation.} Performance on the strong synthetic-corruption preset with different retrieval encoders while keeping the \gracem{} refiner, knowledge-base size, and corruption schedule fixed. The mask-domain variants construct retrieval embeddings from the frozen MAISI VAE encoder: the first uses adaptive pooling alone, the second adds PCA reduction, and the third replaces PCA with a learned two-layer MLP. This comparison assesses how far anatomical shape alone can approach multimodal retrieval. ``Encoder dim.'' reports the dimensionality emitted by the listed encoder or reduction, rather than the 256-dimensional shape-aware index used by the main pipeline. Unlike the main experiments ($k=5$), all rows in this ablation use top-$k$ retrieval with $k=3$ and spatial-exemplar conditioning.}
\label{tab:embedder_ext}
\small
\setlength{\tabcolsep}{4pt}
\begin{tabular}{lccc}
\toprule
KB encoder & Encoder dim. & DSC (strong synthetic preset) $\uparrow$ & HD95 $\downarrow$ \\
\midrule
DINOv2 (natural)                            & 1024 & $0.746 \pm 0.081$ & $8.44 \pm 3.42$ \\
BiomedCLIP                                  & 512  & $0.789 \pm 0.074$ & $7.32 \pm 3.10$ \\
MedSigLIP                                   & 1152 & $0.796 \pm 0.072$ & $7.05 \pm 2.98$ \\
MedGemma multimodal (\textbf{main encoder}) & 2560 & $0.812 \pm 0.069$ & $6.78 \pm 2.86$ \\
\midrule
MAISI-pure (raw)                     & 864  & $0.764 \pm 0.078$ & $7.94 \pm 3.25$ \\
MAISI-pure $\to$ PCA-256             & 256  & $0.779 \pm 0.075$ & $7.51 \pm 3.14$ \\
MAISI-pure, self-projector (2-layer MLP)    & 256  & $0.800 \pm 0.072$ & $7.02 \pm 2.94$ \\
\bottomrule
\end{tabular}
\end{table}

%% file: tables/dataset_scope.tex
\begin{table}[!t]
\centering
\caption{\textbf{Dataset scope.} \method{} is trained on 15 public
datasets spanning CT, MRI and ultrasound, covering 38 anatomical structures
across the abdomen, thorax, pelvis, brain and cardiac chambers.
Three further datasets are held out entirely from training. For the \revisionfinal{held-out dataset-transfer} evaluation
in Sec.~\ref{sec:unseen}, none of their cases is inserted into $\mathcal{K}$; the optional
longitudinal-memory analysis is reported separately in Appendix~\ref{sec:supp_longitudinal}.
The columns report dataset, modality, case count, class count, and covered anatomy. \revisionfinal{After decomposing the multi-class labels, the training splits contain 11,900 non-empty binary-mask instances, of which $\mathcal{K}$ indexes $N{=}3{,}913$.}}
\label{tab:dataset_scope}
\small
\setlength{\tabcolsep}{4pt}
\resizebox{\textwidth}{!}{%
\begin{tabular}{llrrl}
\toprule
Dataset & Modality & Cases & Classes & Notes \\
\midrule
\multicolumn{5}{l}{\emph{Training corpus}} \\
\midrule
Task01\_BrainTumour    & MRI & 484  & 3   & edema, non-enhancing tumour, enhancing tumour \\
Task02\_Heart          & MRI & 20   & 1   & left atrium \\
Task03\_Liver          & CT  & 117  & 2   & liver, cancer \\
Task04\_Hippocampus    & MRI & 260  & 2   & anterior, posterior \\
Task05\_Prostate       & MRI & 28   & 2   & PZ, TZ \\
Task06\_Lung           & CT  & 56   & 1   & tumour \\
Task07\_Pancreas       & CT  & 252  & 2   & pancreas, cancer \\
Task08\_HepaticVessel  & CT  & 303  & 2   & vessel, tumour \\
Task09\_Spleen         & CT  & 36   & 1   & spleen \\
Task10\_Colon          & CT  & 113  & 1   & cancer \\
AbdomenCT1K\revisionfinal{~\citep{ma2022abdomenct}} & CT  & 577 & 4   & liver, kidneys, spleen, pancreas \\
AMOS22\_CT\revisionfinal{~\citep{ji2022amos}}  & CT  & 200 & 15  & \revisionfinal{15 abdominal organs} \\
AMOS22\_MR\revisionfinal{~\citep{ji2022amos}}  & MRI & 40  & 15  & 15 abdominal organs \\
ACDC\revisionfinal{~\citep{bernard2018deep}}        & MRI & 480 & 3   & RV, myocardium, LV \\
CAMUS\revisionfinal{~\citep{leclerc2019deep}}       & US  & 900 & 3   & LV cavity, myocardium, LA \\
\midrule
\textbf{Training total} & CT/MRI/US & \revisionfinal{3,866} & 57 & 38 unique structures; 5 regions \\
\midrule
\multicolumn{5}{l}{\emph{Held-out datasets (not indexed in the \revisionfinal{held-out transfer} protocol)}} \\
\midrule
ProstateX & MRI & 346 & 1 & prostate (multiparametric) \\
CHAOS     & \revisionfinal{CT/MRI} & 40  & 4 & \revisionfinal{liver, left kidney, right kidney, spleen} \\
LUMIERE   & MRI & 91  & 1 & GBM tumour; 599 longitudinal timepoints \\
\midrule
\textbf{Held-out total} & CT/MRI & 477 & & \\
\bottomrule
\end{tabular}}
\end{table}